%% file: paper.tex
\documentclass[sigplan,preprint,authorversion,nonacm]{acmart}
\makeatletter
\@ifundefined{lhs2tex.lhs2tex.sty.read}%
  {\@namedef{lhs2tex.lhs2tex.sty.read}{}%
   \newcommand\SkipToFmtEnd{}%
   \newcommand\EndFmtInput{}%
   \long\def\SkipToFmtEnd#1\EndFmtInput{}%
  }\SkipToFmtEnd

\newcommand\ReadOnlyOnce[1]{\@ifundefined{#1}{\@namedef{#1}{}}\SkipToFmtEnd}
\usepackage{amstext}

\usepackage{amssymb}
\usepackage{stmaryrd}
\DeclareFontFamily{OT1}{cmtex}{}
\DeclareFontShape{OT1}{cmtex}{m}{n}
  {<5><6><7><8>cmtex8
   <9>cmtex9
   <10><10.95><12><14.4><17.28><20.74><24.88>cmtex10}{}
\DeclareFontShape{OT1}{cmtex}{m}{it}
  {<-> ssub * cmtt/m/it}{}

\DeclareFontShape{OT1}{cmtt}{bx}{n}
  {<5><6><7><8>cmtt8
   <9>cmbtt9
   <10><10.95><12><14.4><17.28><20.74><24.88>cmbtt10}{}
\DeclareFontShape{OT1}{cmtex}{bx}{n}
  {<-> ssub * cmtt/bx/n}{}

\newcommand{\Conid}[1]{\mathit{#1}}
\newcommand{\Varid}[1]{\mathit{#1}}
\newcommand{\anonymous}{\kern0.06em \vbox{\hrule\@width.5em}}
\newcommand{\plus}{\mathbin{+\!\!\!+}}
\newcommand{\bind}{\mathbin{>\!\!\!>\mkern-6.7mu=}}

\renewcommand{\geq}{\geqslant}
\usepackage{polytable}

\@ifundefined{mathindent}%
  {\newdimen\mathindent\mathindent\leftmargini}%
  {}%

\def\resethooks{%
  \global\let\SaveRestoreHook\empty
  \global\let\ColumnHook\empty}
\newcommand*{\savecolumns}[1][default]%
  {\g@addto@macro\SaveRestoreHook{\savecolumns[#1]}}
\newcommand*{\restorecolumns}[1][default]%
  {\g@addto@macro\SaveRestoreHook{\restorecolumns[#1]}}
\newcommand*{\aligncolumn}[2]%
  {\g@addto@macro\ColumnHook{\column{#1}{#2}}}

\resethooks

\newcommand{\onelinecommentchars}{\quad-{}- }
\newcommand{\commentbeginchars}{\enskip\{-}
\newcommand{\commentendchars}{-\}\enskip}

\newcommand{\visiblecomments}{%
  \let\onelinecomment=\onelinecommentchars
  \let\commentbegin=\commentbeginchars
  \let\commentend=\commentendchars}

\newcommand{\invisiblecomments}{%
  \let\onelinecomment=\empty
  \let\commentbegin=\empty
  \let\commentend=\empty}

\visiblecomments

\newlength{\blanklineskip}
\newcommand{\hsindent}[1]{\quad}
\let\hspre\empty
\let\hspost\empty

\EndFmtInput
\makeatother
\ReadOnlyOnce{polycode.fmt}%
\makeatletter

\newcommand{\hsnewpar}[1]%
  {{\parskip=0pt\parindent=0pt\par\vskip #1\noindent}}

\newcommand{\hscodestyle}{}

\newcommand{\sethscode}[1]%
  {\expandafter\let\expandafter\hscode\csname #1\endcsname
   \expandafter\let\expandafter\endhscode\csname end#1\endcsname}

  {\par\noindent
   \advance\leftskip\mathindent
   \hscodestyle
   \let\\=\@normalcr
   \let\hspre\(\let\hspost\)%
   \pboxed}%
  {\endpboxed\)%
   \par\noindent
   \ignorespacesafterend}

  {\hsnewpar\abovedisplayskip
   \advance\leftskip\mathindent
   \hscodestyle
   \let\hspre\(\let\hspost\)%
   \pboxed}%
  {\endpboxed%
   \hsnewpar\belowdisplayskip
   \ignorespacesafterend}

  {\hsnewpar\abovedisplayskip
   \advance\leftskip\mathindent
   \hscodestyle
   \let\\=\@normalcr
   \(\pboxed}%
  {\endpboxed\)%
   \hsnewpar\belowdisplayskip
   \ignorespacesafterend}

\newcommand{\plainhs}{\sethscode{plainhscode}}

\plainhs

  {\hsnewpar\abovedisplayskip
   \advance\leftskip\mathindent
   \hscodestyle
   \let\\=\@normalcr
   \(\parray}%
  {\endparray\)%
   \hsnewpar\belowdisplayskip
   \ignorespacesafterend}

  {\parray}{\endparray}

  {\(\parray}{\endparray\)}

\def\codeframewidth{\arrayrulewidth}
\RequirePackage{calc}

  {\parskip=\abovedisplayskip\par\noindent
   \hscodestyle
   \arrayrulewidth=\codeframewidth
   \tabular{@{}|p{\linewidth-2\arraycolsep-2\arrayrulewidth-2pt}|@{}}%
   \hline\framedhslinecorrect\\{-1.5ex}%
   \let\endoflinesave=\\
   \let\\=\@normalcr
   \(\pboxed}%
  {\endpboxed\)%
   \framedhslinecorrect\endoflinesave{.5ex}\hline
   \endtabular
   \parskip=\belowdisplayskip\par\noindent
   \ignorespacesafterend}

\newcommand{\framedhslinecorrect}[2]%
  {#1[#2]}

  {\(\def\column##1##2{}%
   \let\>\undefined\let\<\undefined\let\\\undefined
   \newcommand\>[1][]{}\newcommand\<[1][]{}\newcommand\\[1][]{}%
   \def\fromto##1##2##3{##3}%
   }{\) }%

  {\let\orighscode=\hscode
   \let\origendhscode=\endhscode
   \def\endhscode{\def\hscode{\endgroup\def\@currenvir{hscode}\\}\begingroup}
   \orighscode\def\hscode{\endgroup\def\@currenvir{hscode}}}%
  {\origendhscode
   \global\let\hscode=\orighscode
   \global\let\endhscode=\origendhscode}%

\makeatother
\EndFmtInput

\usepackage[]{algorithmicx}
\usepackage[]{algpseudocode}
\usepackage[]{algorithm}
\usepackage[]{mathpartir}
\usepackage{amsmath}
\usepackage[normalem]{ulem} 

\newcommand{\fungll}[0]{FUN-GLL}
\input{abbrevs}

\begin{document}

\title{{Happy-GLL}: modular, reusable and complete top-down parsers for parameterized nonterminals}

\author{L. Thomas van Binsbergen}
\email{ltvanbinsbergen@acm.org}
\orcid{0000-0001-8113-2221}
\affiliation{
  \institution{Informatics Institute, University of Amsterdam}
  \city{Amsterdam}
  \country{The Netherlands}
}

\author{Damian Fr\"olich}
\email{dfrolich@acm.org}
\orcid{0000-0003-1016-5303}
\affiliation{
  \institution{Informatics Institute, University of Amsterdam}
  \city{Amsterdam}
  \country{The Netherlands}
}


\begin{abstract}
\input{abstract}
\end{abstract}

\maketitle

\renewcommand{\hscodestyle}{\small}

\input{introduction}
\section{Recursive descent and reuse}
\label{sec_motivation}

This section demonstrates the benefits of recursive descent parsing by presenting a translation from Happy grammars to `recognizer\footnote{Recognizers merely indicate whether strings are part of a language whereas parsers also provide a proof in the form of a parse tree.} functions' and motivates parameterized nonterminals by giving examples of reuse.
Familiarity with parsing and Happy (or Yacc-like) syntax is assumed.
The syntax of Happy is explained in the online documentation by~\cite{happy} and~\cite{GruneJacobsParsingTechniques} provide an excellent introduction to parsing.
The Happy grammar in Figure~\ref{fig:syntax_tuple_no_reuse} defines the syntax of a tuple-like structure with alphabetical characters as elements.
Semantic actions, normally associated with production alternates in Happy grammars, are ignored in this section.
The translation generates (possibly higher-order) recognizer functions for the symbols of a grammar (such as \ensuremath{\Varid{alpha}} and \ensuremath{\Conid{AlphaTuples}} in the example).
The recognizer functions may apply the recognizer functions generated for other symbols, independent of whether they are defined in the same file.
Symbol definitions can thus be spread across source files and only need to be recompiled if their definitions change and crucially not when the symbols used in this definition change.
As is described in this section, this improvement over the current (LALR and GLR) back-ends of Happy is achieved by generating top-down parsers and using higher-order functions to implement parameterized nonterminals.
\begin{figure}[t]
\begin{hscode}\SaveRestoreHook
\column{B}{@{}>{\hspre}l<{\hspost}@{}}%
\column{9}{@{}>{\hspre}l<{\hspost}@{}}%
\column{15}{@{}>{\hspre}l<{\hspost}@{}}%
\column{18}{@{}>{\hspre}l<{\hspost}@{}}%
\column{19}{@{}>{\hspre}l<{\hspost}@{}}%
\column{E}{@{}>{\hspre}l<{\hspost}@{}}%
\>[B]{}\%\textbf{name}\;\Varid{tuples}\;{}\<[15]%
\>[15]{}\Conid{AlphaTuples}{}\<[E]%
\\
\>[B]{}\%\textbf{tokentype}\;{}\<[15]%
\>[15]{}\{\mskip1.5mu \Conid{Char}\mskip1.5mu\}{}\<[E]%
\\
\>[B]{}\%\textbf{token}\;{}\<[9]%
\>[9]{}\text{\tt ','}\;{}\<[15]%
\>[15]{}\{\mskip1.5mu \text{\tt ','}\mskip1.5mu\}\;{}\<[E]%
\\
\>[9]{}\text{\tt '('}\;{}\<[15]%
\>[15]{}\{\mskip1.5mu \text{\tt '('}\mskip1.5mu\}\;{}\<[E]%
\\
\>[9]{}\text{\tt ')'}\;{}\<[15]%
\>[15]{}\{\mskip1.5mu \text{\tt ')'}\mskip1.5mu\}{}\<[E]%
\\
\>[B]{}\%\textbf{token}\;\Varid{alpha}\;{}\<[15]%
\>[15]{}\{\mskip1.5mu \Varid{c}\mid \Varid{c}\in ([\mskip1.5mu \text{\tt 'a'}\mathinner{\ldotp\ldotp}\text{\tt 'z'}\mskip1.5mu]\plus [\mskip1.5mu \text{\tt 'A'}\mathinner{\ldotp\ldotp}\text{\tt 'Z'}\mskip1.5mu])\mskip1.5mu\}{}\<[E]%
\\
\>[B]{}\%\%{}\<[E]%
\\
\>[B]{}\Conid{AlphaTuples}{}\<[15]%
\>[15]{}\mathbin{:}{}\<[18]%
\>[18]{}\text{\tt '('}\;\Conid{MAlphas}\;\text{\tt ')'}{}\<[E]%
\\
\>[B]{}\Conid{MAlphas}{}\<[15]%
\>[15]{}\mathbin{:}{}\<[E]%
\\
\>[15]{}\mid {}\<[19]%
\>[19]{}\Conid{Alphas}{}\<[E]%
\\
\>[B]{}\Conid{Alphas}{}\<[15]%
\>[15]{}\mathbin{:}{}\<[19]%
\>[19]{}\Varid{alpha}{}\<[E]%
\\
\>[15]{}\mid {}\<[19]%
\>[19]{}\Varid{alpha}\;\text{\tt ','}\;\Conid{Alphas}{}\<[E]%
\ColumnHook
\end{hscode}\resethooks
\caption{Happy grammar of comma-separated alphabetical characters within parentheses.}
\label{fig:syntax_tuple_no_reuse}
\end{figure}

\subsection{Recognizer functions}
A recognizer function is a function that takes a sequence of tokens \ensuremath{\Varid{inp}} (referred to as a sentence) and an index into the sequence \ensuremath{\Varid{k}} and decides whether it recognizes a prefix of the subsequence of \ensuremath{\Varid{inp}} starting at index \ensuremath{\Varid{k}}.
The type of tokens is left abstract and is denoted in Haskell code by the type variable \ensuremath{\Varid{t}}.
This type is instantiated by the recognizers when there is a \ensuremath{\%\textbf{tokentype}} directive in the grammar specification.
\begin{hscode}\SaveRestoreHook
\column{B}{@{}>{\hspre}l<{\hspost}@{}}%
\column{E}{@{}>{\hspre}l<{\hspost}@{}}%
\>[B]{}\mathbf{type}\;\Conid{Recognizer}\;\Varid{t}\mathrel{=}[\mskip1.5mu \Varid{t}\mskip1.5mu]\to \Conid{Int}\to (\Conid{Int}\to \Conid{Bool})\to \Conid{Bool}{}\<[E]%
\ColumnHook
\end{hscode}\resethooks
The third argument is a continuation function which is to be applied when a prefix is recognized in order to attempt to recognize the rest of the input sentence.
The application of continuation-passing style is crucial to the extension to GLL in Section~\ref{sec_fungll}.
Explanations and applications of continuation-passing for parsing can be found in~\cite{johnson1995,ljunglof2002,swierstra2009,scott2019}.

The \ensuremath{\%\textbf{token}} directives associate an identifier (e.g. \ensuremath{\Varid{alpha}}) with a Haskell pattern written within braces (e.g. \ensuremath{\text{\tt ','}}).
The following recognizer implements token \ensuremath{\Varid{alpha}}:
\begin{hscode}\SaveRestoreHook
\column{B}{@{}>{\hspre}l<{\hspost}@{}}%
\column{3}{@{}>{\hspre}l<{\hspost}@{}}%
\column{10}{@{}>{\hspre}l<{\hspost}@{}}%
\column{13}{@{}>{\hspre}l<{\hspost}@{}}%
\column{23}{@{}>{\hspre}l<{\hspost}@{}}%
\column{55}{@{}>{\hspre}l<{\hspost}@{}}%
\column{E}{@{}>{\hspre}l<{\hspost}@{}}%
\>[B]{}\mbox{\onelinecomment  \ensuremath{\Varid{t}} instantiated to Char following the $\%tokentype$ declaration}{}\<[E]%
\\
\>[B]{}p\_\mathit{alpha}\mathbin{::}\Conid{Recognizer}\;\Conid{Char}{}\<[E]%
\\
\>[B]{}p\_\mathit{alpha}\mathrel{=}\Varid{matchPattern\char95 1}\;\Varid{matcher}{}\<[E]%
\\
\>[B]{}\hsindent{3}{}\<[3]%
\>[3]{}\mathbf{where}\;{}\<[10]%
\>[10]{}\Varid{matcher}\;\Varid{t}\mathrel{=}{}\<[23]%
\>[23]{}\mathbf{case}\;\Varid{t}\;\mathbf{of}{}\<[E]%
\\
\>[10]{}\hsindent{3}{}\<[13]%
\>[13]{}\Varid{c}\mid \Varid{c}\in ([\mskip1.5mu \text{\tt 'a'}\mathinner{\ldotp\ldotp}\text{\tt 'z'}\mskip1.5mu]\plus [\mskip1.5mu \text{\tt 'A'}\mathinner{\ldotp\ldotp}\text{\tt 'Z'}\mskip1.5mu]){}\<[55]%
\>[55]{}\to \Conid{Just}\;\Varid{t}{}\<[E]%
\\
\>[10]{}\hsindent{3}{}\<[13]%
\>[13]{}\anonymous {}\<[55]%
\>[55]{}\to \Conid{Nothing}{}\<[E]%
\ColumnHook
\end{hscode}\resethooks
The \ensuremath{\%\textbf{token}} directive determines the recognizer's name and the first case of the \ensuremath{\mathbf{case}}-expression.
The logic of recognizer functions for tokens is implemented in the function \ensuremath{\Varid{matchPattern\char95 1}}, provided in a separate support library, as follows:
\begin{hscode}\SaveRestoreHook
\column{B}{@{}>{\hspre}l<{\hspost}@{}}%
\column{3}{@{}>{\hspre}c<{\hspost}@{}}%
\column{3E}{@{}l@{}}%
\column{7}{@{}>{\hspre}l<{\hspost}@{}}%
\column{61}{@{}>{\hspre}l<{\hspost}@{}}%
\column{E}{@{}>{\hspre}l<{\hspost}@{}}%
\>[B]{}\Varid{matchPattern\char95 1}\mathbin{::}(\Varid{t}\to \Conid{Maybe}\;\Varid{t})\to \Conid{Recognizer}\;\Varid{t}{}\<[E]%
\\
\>[B]{}\Varid{matchPattern\char95 1}\;\Varid{matcher}\;\Varid{inp}\;\Varid{k}\;\Varid{c}{}\<[E]%
\\
\>[B]{}\hsindent{3}{}\<[3]%
\>[3]{}\mid {}\<[3E]%
\>[7]{}\Varid{k}\geq \mathrm{0},\Varid{k}\mathbin{<}\Varid{length}\;\Varid{inp},\Conid{Just}\;\anonymous \leftarrow \Varid{matcher}\;(\Varid{inp}\mathbin{!!}\Varid{k}){}\<[61]%
\>[61]{}\mathrel{=}\Varid{c}\;(\Varid{k}\mathbin{+}\mathrm{1}){}\<[E]%
\\
\>[B]{}\hsindent{3}{}\<[3]%
\>[3]{}\mid {}\<[3E]%
\>[7]{}\Varid{otherwise}{}\<[61]%
\>[61]{}\mathrel{=}\Conid{False}{}\<[E]%
\ColumnHook
\end{hscode}\resethooks
The continuation \ensuremath{\Varid{c}} is applied (to \ensuremath{\Varid{k}\mathbin{+}\mathrm{1}}, because one token has been recognized) when the token is successfully matched and \ensuremath{\Conid{False}} is returned otherwise.

A recognizer function can be given to the support function \ensuremath{\Varid{run}_1} to attempt to recognize a sentence.
The function applies the recognizer to the sentence, together with an initial index of $0$ and a continuation that checks whether the whole sentence has been recognized.
\begin{hscode}\SaveRestoreHook
\column{B}{@{}>{\hspre}l<{\hspost}@{}}%
\column{E}{@{}>{\hspre}l<{\hspost}@{}}%
\>[B]{}\Varid{run}_1\mathbin{::}\Conid{Recognizer}\;\Varid{t}\to [\mskip1.5mu \Varid{t}\mskip1.5mu]\to \Conid{Bool}{}\<[E]%
\\
\>[B]{}\Varid{run}_1\;\Varid{rec}\;\Varid{inp}\mathrel{=}\Varid{rec}\;\Varid{inp}\;\mathrm{0}\;(\lambda \Varid{k}\to \Varid{k}\equiv \Varid{length}\;\Varid{inp}){}\<[E]%
\ColumnHook
\end{hscode}\resethooks

The function \ensuremath{\Varid{run}_1} is used to implement the \ensuremath{\%\textbf{name}} directive of Happy (indicating the start symbol to be used by the LR back-ends).
The following code is generated for the \ensuremath{\%\textbf{name}} directive of the example.
The directive determines the name of the function and the used recognizer.
\begin{hscode}\SaveRestoreHook
\column{B}{@{}>{\hspre}l<{\hspost}@{}}%
\column{E}{@{}>{\hspre}l<{\hspost}@{}}%
\>[B]{}\mathit{run}\_\mathit{tuples}\mathbin{::}[\mskip1.5mu \Varid{t}\mskip1.5mu]\to \Conid{Bool}{}\<[E]%
\\
\>[B]{}\mathit{run}\_\mathit{tuples}\mathrel{=}\Varid{run}_1\;p\_\mathit{AlphaTuples}{}\<[E]%
\ColumnHook
\end{hscode}\resethooks
The application of \ensuremath{\Varid{run}_1} demonstrates one of the advantages of our approach: every recognizer can be used to recognize a sentence, without the need for recompilation.
%

To generate code for a nonterminal's alternates it is required to place recognizers in sequence so that one is executed `before' the other. 
For example, the last alternate of \ensuremath{\Conid{Alphas}} requires an \ensuremath{\Varid{alpha}} to be recognized, then a comma (\ensuremath{\text{\tt ','}}) and then more \ensuremath{\Conid{Alphas}}.
For every alternate, code is generated that `chains' several recognizer functions by defining new continuations.
For example, if symbol \ensuremath{\Varid{p}} appears before \ensuremath{\Varid{q}} in the alternate of a nonterminal, then the recognizer for \ensuremath{\Varid{p}} is applied with a continuation that applies the recognizer for \ensuremath{\Varid{q}}.
The following code is generated for nonterminal \ensuremath{\Conid{Alphas}}:
\begin{hscode}\SaveRestoreHook
\column{B}{@{}>{\hspre}l<{\hspost}@{}}%
\column{3}{@{}>{\hspre}l<{\hspost}@{}}%
\column{11}{@{}>{\hspre}l<{\hspost}@{}}%
\column{E}{@{}>{\hspre}l<{\hspost}@{}}%
\>[B]{}p\_\mathit{Alphas}\mathbin{::}\Conid{Recognizer}\;\Conid{Char}{}\<[E]%
\\
\>[B]{}p\_\mathit{Alphas}\;\Varid{inp}\;\Varid{k}\;c_0\mathrel{=}p\_\mathit{alpha}\;\Varid{inp}\;\Varid{k}\;c_0\mathrel{\vee}p\_\mathit{alpha}\;\Varid{inp}\;\Varid{k}\;c_1{}\<[E]%
\\
\>[B]{}\hsindent{3}{}\<[3]%
\>[3]{}\mathbf{where}\;{}\<[11]%
\>[11]{}c_1\;\Varid{k}\mathrel{=}p\_\mathit{comma}\;\Varid{inp}\;\Varid{k}\;c_2{}\<[E]%
\\
\>[11]{}c_2\;\Varid{k}\mathrel{=}p\_\mathit{Alphas}\;\Varid{inp}\;\Varid{k}\;c_0{}\<[E]%
\ColumnHook
\end{hscode}\resethooks
The code for both alternates of the nonterminal defining \ensuremath{\Conid{Alphas}} apply \ensuremath{p\_\mathit{alpha}}.
The difference is in the continuation given to \ensuremath{p\_\mathit{alpha}}.
The first alternate is done after \ensuremath{p\_\mathit{alpha}} so the given continuation (\ensuremath{c_0}) is the continuation received by \ensuremath{p\_\mathit{Alphas}}.
In the second alternate, \ensuremath{\Varid{alpha}} is followed by \ensuremath{\Varid{comma}} and \ensuremath{\Conid{Alphas}} and thus\footnote{The token name \ensuremath{\text{\tt ','}} given to the pattern \ensuremath{\text{\tt ','}} by the first \ensuremath{\%\textbf{token}} directive is replaced by \ensuremath{\Varid{comma}} in the name for the generated recognizer.} \ensuremath{c_1\;\Varid{k}\mathrel{=}p\_\mathit{comma}\;\Varid{inp}\;\Varid{k}\;c_2} and \ensuremath{c_2\;\Varid{k}\mathrel{=}p\_\mathit{Alphas}\;\Varid{inp}\;\Varid{k}\;c_0}.
Note that these definitions have shadowing declarations for the variable \ensuremath{\Varid{k}} in order to keep a consistent naming convention.

The first alternate of \ensuremath{\Conid{MAlphas}} is empty. 
The code for this alternate therefore directly applies the continuation $c_0$:
\begin{hscode}\SaveRestoreHook
\column{B}{@{}>{\hspre}l<{\hspost}@{}}%
\column{E}{@{}>{\hspre}l<{\hspost}@{}}%
\>[B]{}p\_\mathit{MAlphas}\mathbin{::}\Conid{Recognizer}\;\Conid{Char}{}\<[E]%
\\
\>[B]{}p\_\mathit{MAlphas}\;\Varid{inp}\;\Varid{k}\;c_0\mathrel{=}c_0\;\Varid{k}\mathrel{\vee}p\_\mathit{Alphas}\;\Varid{inp}\;\Varid{k}\;c_0{}\<[E]%
\ColumnHook
\end{hscode}\resethooks
The recognizer\footnote{The names for \ensuremath{\text{\tt '('}} and \ensuremath{\text{\tt ')'}} have been replaced by \ensuremath{\Varid{lparen}} and \ensuremath{\Varid{rparen}}.} for \ensuremath{\Conid{AlphaTuples}} completes the translation of the grammar in Figure~\ref{fig:syntax_tuple_no_reuse}:
\begin{hscode}\SaveRestoreHook
\column{B}{@{}>{\hspre}l<{\hspost}@{}}%
\column{13}{@{}>{\hspre}l<{\hspost}@{}}%
\column{21}{@{}>{\hspre}l<{\hspost}@{}}%
\column{E}{@{}>{\hspre}l<{\hspost}@{}}%
\>[B]{}p\_\mathit{AlphaTuples}\mathbin{::}\Conid{Recognizer}\;\Conid{Char}{}\<[E]%
\\
\>[B]{}p\_\mathit{AlphaTuples}\;\Varid{inp}\;\Varid{k}\;c_0\mathrel{=}p\_\mathit{lparen}\;\Varid{inp}\;\Varid{k}\;c_1{}\<[E]%
\\
\>[B]{}\hsindent{13}{}\<[13]%
\>[13]{}\mathbf{where}\;{}\<[21]%
\>[21]{}c_1\;\Varid{k}\mathrel{=}p\_\mathit{MAlphas}\;\Varid{inp}\;\Varid{k}\;c_2{}\<[E]%
\\
\>[21]{}c_2\;\Varid{k}\mathrel{=}p\_\mathit{rparen}\;\Varid{inp}\;\Varid{k}\;c_0{}\<[E]%
\ColumnHook
\end{hscode}\resethooks
\subsection{Discussion}
\paragraph{Reuse}
The parameterized nonterminals of Happy abstract over symbols.
For example, \ensuremath{\Varid{comma}} in the definition of \ensuremath{\Conid{Alphas}} can be replaced by a parameter to abstract over \ensuremath{\Varid{comma}} as a separator.
Abstractions and applications are written in the familiar `functional style', with formal and actual parameters appearing between parentheses and with commas in between.
A parameterized nonterminal for the suggested, more general version of \ensuremath{\Conid{Alphas}} is written as follows:
\begin{hscode}\SaveRestoreHook
\column{B}{@{}>{\hspre}l<{\hspost}@{}}%
\column{17}{@{}>{\hspre}c<{\hspost}@{}}%
\column{17E}{@{}l@{}}%
\column{20}{@{}>{\hspre}l<{\hspost}@{}}%
\column{E}{@{}>{\hspre}l<{\hspost}@{}}%
\>[B]{}\Conid{SepAlphas}\;(\Varid{sep}){}\<[17]%
\>[17]{}\mathbin{:}{}\<[17E]%
\>[20]{}\Varid{alpha}{}\<[E]%
\\
\>[17]{}\mid {}\<[17E]%
\>[20]{}\Varid{alpha}\;\Varid{sep}\;\Conid{SepAlphas}\;(\Varid{sep}){}\<[E]%
\ColumnHook
\end{hscode}\resethooks

By taking advantage of Haskell's higher-order functions, the abstractions and applications of Happy can be translated more or less directly into Haskell:
\begin{hscode}\SaveRestoreHook
\column{B}{@{}>{\hspre}l<{\hspost}@{}}%
\column{10}{@{}>{\hspre}l<{\hspost}@{}}%
\column{E}{@{}>{\hspre}l<{\hspost}@{}}%
\>[B]{}p\_\mathit{SepAlphas}\mathbin{::}\Conid{Recognizer}\;\Conid{Char}\to \Conid{Recognizer}\;\Conid{Char}{}\<[E]%
\\
\>[B]{}p\_\mathit{SepAlphas}\;p\_\mathit{sep}\;\Varid{inp}\;\Varid{k}\;c_0\mathrel{=}p\_\mathit{alpha}\;\Varid{inp}\;\Varid{k}\;c_0\mathrel{\vee}p\_\mathit{alpha}\;\Varid{inp}\;\Varid{k}\;c_1{}\<[E]%
\\
\>[B]{}\mathbf{where}\;{}\<[10]%
\>[10]{}c_1\;\Varid{k}\mathrel{=}p\_\mathit{sep}\;\Varid{inp}\;\Varid{k}\;c_2{}\<[E]%
\\
\>[10]{}c_2\;\Varid{k}\mathrel{=}p\_\mathit{Alphas}\;p\_\mathit{sep}\;\Varid{inp}\;\Varid{k}\;c_0{}\<[E]%
\ColumnHook
\end{hscode}\resethooks
\begin{figure}[t]
\begin{hscode}\SaveRestoreHook
\column{B}{@{}>{\hspre}l<{\hspost}@{}}%
\column{14}{@{}>{\hspre}l<{\hspost}@{}}%
\column{16}{@{}>{\hspre}c<{\hspost}@{}}%
\column{16E}{@{}l@{}}%
\column{17}{@{}>{\hspre}l<{\hspost}@{}}%
\column{20}{@{}>{\hspre}l<{\hspost}@{}}%
\column{21}{@{}>{\hspre}c<{\hspost}@{}}%
\column{21E}{@{}l@{}}%
\column{25}{@{}>{\hspre}l<{\hspost}@{}}%
\column{E}{@{}>{\hspre}l<{\hspost}@{}}%
\>[B]{}\Conid{Tuples}\;(\Varid{elem}){}\<[17]%
\>[17]{}\mathbin{:}\Conid{Parens}\;(\Conid{Optional}\;(\Conid{Multiple}\;(\Varid{elem},\Varid{comma}))){}\<[E]%
\\
\>[B]{}\Conid{Lists}\;(\Varid{elem}){}\<[14]%
\>[14]{}\mathbin{:}\Conid{Brackets}\;(\Conid{Optional}\;(\Conid{Multiple}\;(\Varid{elem},\Varid{comma}))){}\<[E]%
\\[\blanklineskip]%
\>[B]{}\Conid{Parens}\;(\Varid{x}){}\<[16]%
\>[16]{}\mathbin{:}{}\<[16E]%
\>[20]{}\Conid{Within}\;(\text{\tt '('},\text{\tt ')'},\Varid{x}){}\<[E]%
\\
\>[B]{}\Conid{Brackets}\;(\Varid{x}){}\<[16]%
\>[16]{}\mid {}\<[16E]%
\>[20]{}\Conid{Within}\;(\text{\tt '['},\text{\tt ']'},\Varid{x}){}\<[E]%
\\
\>[B]{}\Conid{Within}\;(\Varid{l},\Varid{r},\Varid{x}){}\<[16]%
\>[16]{}\mathbin{:}{}\<[16E]%
\>[20]{}\Varid{l}\;\Varid{x}\;\Varid{r}{}\<[E]%
\\[\blanklineskip]%
\>[B]{}\Conid{Optional}\;(\Varid{x}){}\<[21]%
\>[21]{}\mathbin{:}{}\<[21E]%
\\
\>[21]{}\mid {}\<[21E]%
\>[25]{}\Varid{x}{}\<[E]%
\\[\blanklineskip]%
\>[B]{}\Conid{Multiple}\;(\Varid{elem},\Varid{sep}){}\<[21]%
\>[21]{}\mathbin{:}{}\<[21E]%
\>[25]{}\Varid{elem}{}\<[E]%
\\
\>[21]{}\mid {}\<[21E]%
\>[25]{}\Varid{elem}\;\Varid{sep}\;\Conid{Multiple}\;(\Varid{elem},\Varid{sep}){}\<[E]%
\ColumnHook
\end{hscode}\resethooks
\caption{Examples of reusable parameterized nonterminals.}%
\label{fig:syntax_tuple_reuse}%
\end{figure}
\indent Figure~\ref{fig:syntax_tuple_reuse} shows several parameterized nonterminals for highly reusable patterns such as delimiters, repetition with a separator, and optionality, as well as examples of their usage.
%
%
Further real-world examples for reuse can be found in~\cite{thiemann2008}.

\paragraph{Advantages of recursive descent}
LALR and GLR require\footnote{The \ensuremath{\%\textbf{name}} directive is optional though, because by default the first nonterminal of the file is chosen as the entry point.} the indication of `entry point nonterminals' via \ensuremath{\%\textbf{name}} directives.
This is because LR algorithms require grammars with start symbols.
%
%
The translation described in this section does not require entry points because \ensuremath{\Varid{run}_1} can be applied to arbitrary recognizers.
Practically, this means that every nonterminal of a grammar can be debugged individually, making the whole grammar easier to test and maintain.
Moreover, the code generated for a nonterminal is independent of the definitions of the other symbols in the grammar.
The translation thus preserves the inherent modularity of nonterminals.

This section has shown that recognizer functions are freely combined to form more complex recognizers in higher-order functions such as \ensuremath{p\_\mathit{SepAlphas}}.
Recognizers generated from different grammars can also be combined in this way, enabling reuse across files, albeit with the following caveats.
Type signatures instantiate the type for tokens to the type mentioned in the \ensuremath{\%\textbf{tokentype}} directive.
This implies that two grammars with different \ensuremath{\%\textbf{tokentype}} directives produces recognizers that cannot be composed.
There are at least two ways a Happy user can approach this potential problem, depending on the application. 

One approach is to omit the \ensuremath{\%\textbf{tokentype}} directive and to rely on Haskell's type-inferencing to automatically assign the most general type possible.
If the \ensuremath{\%\textbf{tokentype}} directive is removed from the grammar in Figure~\ref{fig:syntax_tuple_no_reuse}, then the inferred type is \ensuremath{\Conid{Char}} because of the usage of literals in the \ensuremath{\%\textbf{token}} directives.
Although the definition of \ensuremath{\Conid{MAlphas}} does not refer to token symbols, the token type \ensuremath{\Conid{Char}} is inferred from the type of the recognizer for \ensuremath{\Conid{Alphas}}.
If \ensuremath{\Conid{Alphas}} was a parameter -- as it is in \ensuremath{\Conid{Optional}} of Figure~\ref{fig:syntax_tuple_reuse} -- then the derived type would have been a type variable.
The function and signature generated for \ensuremath{\Conid{Optional}} are as follows\footnotemark{}:
\begin{hscode}\SaveRestoreHook
\column{B}{@{}>{\hspre}l<{\hspost}@{}}%
\column{E}{@{}>{\hspre}l<{\hspost}@{}}%
\>[B]{}p\_\mathit{Optional}\mathbin{::}\Conid{Recognizer}\;\Varid{t}\to \Conid{Recognizer}\;\Varid{t}{}\<[E]%
\\
\>[B]{}p\_\mathit{Optional}\;\Varid{p\char95 x}\;\Varid{inp}\;\Varid{k}\;c_0\mathrel{=}c_0\;\Varid{k}\mathrel{\vee}\Varid{p\char95 x}\;\Varid{inp}\;\Varid{k}\;c_0{}\<[E]%
\ColumnHook
\end{hscode}\resethooks
\footnotetext{%
The translation by the GLL back-end differs slightly because type inferencing might assign constraints to types.
%
%
Rather than inferring types, the GLL back-end uses the `wildcards' of the \texttt{PartialTypeSignatures} extension of Haskell.
The wildcards are replaced with (appropriately constrained) types by compilers supporting this extension.
%
%
}%
\indent Another approach to reusing nonterminal definitions across grammars is to give co-dependent grammars the same type for tokens.
This might be practical for projects that involve a single lexer or that have multiple lexers producing tokens of the same type.
This approach provides a practical improvement over the LALR and GLR back-ends because large, complex grammars, with perhaps many entry points, can be spread across several files.
Moreover, adopting this approach does not rule out defining a reusable library of parameterized nonterminals (such as the ones in Figure~\ref{fig:syntax_tuple_reuse}) in a separate file without \ensuremath{\%\textbf{tokentype}} directive.
Note that the alternates for \ensuremath{\Conid{Within}}, \ensuremath{\Conid{Optional}} and \ensuremath{\Conid{Multiple}} in Figure~\ref{fig:syntax_tuple_reuse} do not refer to any token symbols.

\paragraph{Limitations of basic recursive descent}
\label{sec_rdp_limitations}
%
%
%

Recursive descent parsers typically choose an alternate of a nonterminal based on lookahead (the recognizer functions of this section lazily apply all alternates without considering lookahead).
Lookahead involves pre-computing for each alternate the set of terminals the alternate is capable of matching initially, and checking, during a parse, whether the next terminal in the input sentence is a member of that set. 
In general, however, lookahead cannot be used to rule out each, or all but one, alternate.
The class of LL(k) grammars is defined to contain those context-free grammars for which it holds that no two alternates are simultaneously applicable using $k \geq 0$ symbols of lookahead.
After choosing an alternate, different forms of backtracking can be used to revert the decision and to choose another alternate.
For every amount of lookahead $k$ and for every backtracking strategy, however, there are worst-case grammars that require running times exponential in the size of the input.
These exponential running times can be avoided with memoization~\cite{norvig1991,frost1996}.

If a nonterminal is left-recursive, a parse function implementing this non-terminal may end up calling itself without a change of input, resulting in non-termination.
Grammar transformations can be used to remove left-recursion and to left-factorize the grammar, turning the grammar into an LL(k) grammar.
However, not every context-free grammar has an LL(k) equivalent and transforming a grammar is not desirable if the resulting grammar is further removed from the semantic interpretation of the syntax.
Memoization tables can be used to record continuations for parse functions when parse functions are written in continuation-passing style, making it possible to produce complete parsers for left-recursive grammars~\cite{johnson1995,afroozeh2016}.
Frost, Hafiz, and Callaghan employ a `curtailment' strategy to handle left-recursion, making at most as many recursive calls as there are characters left in the sentence~\cite{frost2008}.

The parsers generated by the GLL back-end of Happy manage continuations in a fashion similar to~\cite{johnson1995,afroozeh2016} to avoid all forms of repeated work -- thereby preventing non-termination and exponential running times -- and to discover all possible derivations.
The grammar to GLL parser translation, presented in the next section, also maintains the benefits of recursive descent parsing mentioned in this section.

\section{Generating {GLL} parsers}
\label{sec_fungll}
This section explains how the GLL back-end translates Happy grammars to GLL parsers based on the purely functional (\fungll{}) variant of GLL presented here~\cite{binsbergen2018a,binsbergen2020}.
In~\cite{binsbergen2020}, it is shown that \fungll{} can be implemented directly as parser combinators without generating a grammar object as in~\cite{binsbergen2018a}.
The insights that enabled the parser combinators of~\cite{binsbergen2020} also enable the translation presented in this section. 
The \fungll{} algorithm and its combinator-oriented origins in~\cite{binsbergen2018a} and~\cite{binsbergen2020} are summarized first.

The combinator expressions written with the parser combinators of~\cite{binsbergen2018a} evaluate in three stages.
First, the grammar represented by the combinator expression is extracted.
The grammar is then given to a standalone implementation of \fungll{} which produces an efficient representation of all possible derivations: a set of BSR elements named after the binary subtree representation (BSR) of~\cite{scott2019}.
An evaluation function is extracted alongside the grammar.
The evaluation function is applied to the BSR set to compute a semantic value for each of the derivations by executing the semantic actions of the combinator expression (this step is hereafter referred to as the `semantic phase').
A similar architecture for developing complete parsers with grammar combinators was first presented by Ridge~\cite{ridge2014}.
In~\cite{binsbergen2020} it was shown that parser combinators can use the datastructures of \fungll{} to compute BSR sets directly, thereby removing the need to compute an intermediate grammar object.
The key insight is that the grammar information used by \fungll{} can be computed locally for each nonterminal.
It is this insight that makes it possible to generate complete, modular and reusable top-down parsers from Happy grammars and to translate Happy's parameterized nonterminals directly.

\subsection{The \fungll{} algorithm}
\newcommand{\gslot}[3]{{#1::=#2\,\raisebox{.15em}{$\scriptscriptstyle\bullet$}\, #3}}
\newcommand{\bsr}[6]{{\langle\gslot{#1}{#2}{#3},#4,#5,#6\rangle}}
\newcommand{\descr}[5]{{\langle\gslot{#1}{#2}{#3},#4,#5\rangle}}
\newcommand{\descrc}{\refstepcounter{descr}\thedescr}
The following paragraphs summarize the explanations of the \fungll{} algorithm in~\cite{binsbergen2018a,binsbergen_thesis,binsbergen2020}.
An example run of the algorithm is given in Table~\ref{tab_run_eee}, with the output shown in Figure~\ref{fig_eee_bsr}, demonstrating how the algorithm deals with left-recursion and ambiguity.
The example is taken from~\cite{binsbergen2020} based on the grammar used by Ridge~\cite{ridge2014}.
\input{example_fungll}

Consider a simple recursive descent parser with a parse function for each symbol of a grammar that receives as argument an index $l$ into the input sentence and returns an index $r \geq l$.
If the parse function for \ensuremath{\Varid{x}} returns $r$ given $l$, then \ensuremath{\Varid{x}} matches the subsentence ranging from $l$ to $r-1$. 
The indices $l$ and $r$ are referred to as the \emph{left extent} and \emph{right extent} of the match respectively.
The \fungll{} algorithm generalizes such recursive descent parsers by trying all alternates of \ensuremath{\Varid{x}} in order to find all right extents.
The algorithm prevents duplicate work, and thereby non-termination and exponential running times, by labelling parts of the work with \emph{descriptors} and ensuring that no descriptor is `processed' twice.
A descriptors is a triple \ensuremath{((\Varid{x},\alpha,\beta),\Varid{l},\Varid{k})} with \ensuremath{(\Varid{x},\alpha,\beta)} a \emph{slot} and $k \geq l$ integers.
A slot \ensuremath{(\Varid{x},\alpha,\beta)} identifies the point in the alternate $\alpha\beta$ of \ensuremath{\Varid{x}} preceded by the symbols \ensuremath{\alpha} and succeeded by the symbols \ensuremath{\beta}.
A descriptor \ensuremath{((\Varid{x},\alpha,\beta),\Varid{l},\Varid{k})} denotes progress towards checking whether the alternate $\alpha\beta$ can be used to match a subsentence with left extent $l$, indicating that the symbols in \ensuremath{\alpha} have matched with left extent \ensuremath{\Varid{l}} and right extent \ensuremath{\Varid{k}}.
At any time during the execution of the algorithm, the set \ensuremath{\Varid{uset}} holds the descriptors encountered so far.

The relation \ensuremath{\Varid{prel}} holds the right extents discovered for a pair \ensuremath{(\Varid{x},\Varid{l})} of a nonterminal and a left extent, referred to as a \emph{commencement} (as a dual to `continuation'). 
If the element \ensuremath{((\Varid{x},\Varid{l}),\Varid{r})} is in \ensuremath{\Varid{prel}}, then \ensuremath{\Varid{r}} is a right extent for the commencement \ensuremath{(\Varid{x},\Varid{l})}.
Relation \ensuremath{\Varid{prel}} is reminiscent of a memoization table for a parse function of \ensuremath{\Varid{x}}.

The relation \ensuremath{\Varid{grel}} associates a commencement with zero or more continuation identifiers.
A continuation identifier \ensuremath{((\Varid{x},\alpha,\beta),\Varid{l})} is essentially a descriptor with a hole for the second index, which is to be filled by every right extent discovered for the sequence of symbols \ensuremath{\alpha}.
If \ensuremath{((\Varid{y},\Varid{k}),((\Varid{x},\alpha,\beta),\Varid{l}))} is in \ensuremath{\Varid{grel}} and if right extent \ensuremath{\Varid{r}} is discovered for the commencement \ensuremath{(\Varid{y},\Varid{k})}, this indicates that \ensuremath{\Varid{r}} is a right extent for \ensuremath{\alpha} and thus that the descriptor \ensuremath{((\Varid{x},\alpha,\beta),\Varid{l},\Varid{r})} needs to be processed (if not in \ensuremath{\Varid{uset}}).
The algorithm is such that if \ensuremath{((\Varid{y},\Varid{k}),((\Varid{x},\alpha,\beta),\Varid{l}))} is in \ensuremath{\Varid{grel}}, then the last symbol of \ensuremath{\alpha} is \ensuremath{\Varid{y}} and there is a descriptor $((x,\alpha',y\beta),l,k)$ in \ensuremath{\Varid{uset}} with $\alpha = \alpha'y$.

The algorithm records that descriptor $((x,\alpha',y\beta),l,k)$ and the right extent $r$ for commencement \ensuremath{(\Varid{y},\Varid{k})} give rise to the descriptor $((x,\alpha'y,\beta),l,r)$ by adding the BSR element $((x,\alpha'y,\beta),l,k,r)$ to the set \ensuremath{\Varid{bsrs}}. 
The index $k$ in some sense connects the two consecutive descriptors $((x,\alpha',y\beta),l,k)$ and $((x,\alpha'y,\beta),l,r)$ and is therefore referred to as a \emph{pivot}.
At any moment, \ensuremath{\Varid{bsrs}} contains enough information to retrace the entire parse up to that moment because \ensuremath{\Varid{bsrs}} records the pivots for every pair of consecutive descriptors.
The reader is referred to~\cite{scott2019} for a full explanation on how a BSR set embeds derivations.
As explained later, the semantic phase revisits all traces of complete matches of the sentence (potentially subjected to ambiguity reduction strategies) and executes semantic actions along the way.
Based on the introduced sets and relations, the \fungll{} algorithm can be explained in terms of the actions \textbf{descend}, \textbf{ascend}, \textbf{process} and \textbf{continue}:
\begin{itemize}
\item
To \textbf{descend} a commencement \ensuremath{(\Varid{y},\Varid{k})} with continuation identifier \ensuremath{((\Varid{x},\alpha,\beta),\Varid{l})} means: extending \ensuremath{\Varid{grel}} to contain \ensuremath{((\Varid{y},\Varid{k}),((\Varid{x},\alpha,\beta),\Varid{l}))} and finding the set $R$ such that for all $r\in R$ it holds that \ensuremath{((\Varid{y},\Varid{k}),\Varid{r})} is in \ensuremath{\Varid{prel}}.
\begin{itemize}
  \item If $R \neq \emptyset$, then \textbf{continue} with descriptor \ensuremath{((\Varid{x},\alpha,\beta),\Varid{l},\Varid{r})} and pivot \ensuremath{\Varid{k}}, for every $r \in R$ and in any order
  \item If $R = \emptyset$, \textbf{process} the descriptor \ensuremath{((\Varid{y},\epsilon,\delta),\Varid{k},\Varid{k})}, for every alternate $\delta$ of $y$ and in any order (\ensuremath{\epsilon} denotes the empty sequence of symbols)
\end{itemize}
\item To \textbf{ascend} a commencement \ensuremath{(\Varid{x},\Varid{k})} with right extent $r$ means: extending \ensuremath{\Varid{prel}} to contain \ensuremath{((\Varid{x},\Varid{k}),\Varid{r})} and finding all continuation identifiers \ensuremath{(\Varid{s},\Varid{l})} for which it holds that \ensuremath{((\Varid{x},\Varid{k}),(\Varid{s},\Varid{l}))} is in \ensuremath{\Varid{grel}} and \textbf{continue} with every descriptors \ensuremath{(\Varid{s},\Varid{l},\Varid{r})} and pivot \ensuremath{\Varid{k}} in any order
\item To \textbf{continue} with a descriptor \ensuremath{((\Varid{x},\alpha,\beta),\Varid{l},\Varid{r})} and pivot \ensuremath{\Varid{k}} means: add the BSR element \ensuremath{((\Varid{x},\alpha,\beta),\Varid{l},\Varid{k},\Varid{r})} to \ensuremath{\Varid{bsrs}} and then \textbf{process} descriptor \ensuremath{((\Varid{x},\alpha,\beta),\Varid{l},\Varid{r})}
\item A descriptor \ensuremath{((\Varid{x},\alpha,\beta),\Varid{l},\Varid{r})} is processed by adding it to \ensuremath{\Varid{uset}} and only if it was not already in \ensuremath{\Varid{uset}} do the following:
\begin{itemize}
\item If \ensuremath{\beta} is the empty sequence of symbols, then \textbf{ascend} commencement \ensuremath{(\Varid{x},\Varid{l})} with right extent $r$. 
\item If $\beta = t\beta'$ with $t$ a token, match $t$ it against the token at position $r$ in the sentence and, if successful, \textbf{continue} with ${((x,\alpha t,\beta'),l,r+1)}$ and pivot \ensuremath{\Varid{r}}
\item If $\beta = y\beta'$ with \ensuremath{\Varid{y}} a nonterminal, \textbf{descend} \ensuremath{(\Varid{y},\Varid{r})} with continuation identifier ${((x,\alpha y,\beta'),l)}$
\end{itemize} 
\end{itemize} 
The example run in Table~\ref{tab_run_eee} starts with processing the descriptors $\descr{E}{}{EEE}{0}{0}$, $\descr{E}{}{\texttt{a}}{0}{0}$ and $\descr{E}{}{}{0}{0}$. 
The symbol $\raisebox{.15em}{$\scriptscriptstyle\bullet$}$ marks the position in an alternate identified by a slot.
Duplicate set elements are \sout{striked} (the action for descriptor \#10 is striked because the character `a' is not the second character in the input).
The descriptors are processed in the order of first-in first out, meaning that the $i$-th descriptor to be encountered is also the $i$-th descriptor to be processed.
Choosing this order has enabled the concise tabular form of presenting runs of the algorithm; any other order gives the same output.

\subsection{Translating Happy grammars}
\begin{figure}[t]%
\begin{hscode}\SaveRestoreHook
\column{B}{@{}>{\hspre}l<{\hspost}@{}}%
\column{14}{@{}>{\hspre}l<{\hspost}@{}}%
\column{17}{@{}>{\hspre}l<{\hspost}@{}}%
\column{23}{@{}>{\hspre}l<{\hspost}@{}}%
\column{24}{@{}>{\hspre}l<{\hspost}@{}}%
\column{29}{@{}>{\hspre}l<{\hspost}@{}}%
\column{32}{@{}>{\hspre}l<{\hspost}@{}}%
\column{34}{@{}>{\hspre}l<{\hspost}@{}}%
\column{E}{@{}>{\hspre}l<{\hspost}@{}}%
\>[B]{}\mathbf{data}\;\Conid{SID}{}\<[14]%
\>[14]{}\mathrel{=}\Conid{Tn}\;\Conid{String}\mid \Conid{App}\;\Conid{String}\;[\mskip1.5mu \Conid{SID}\mskip1.5mu]{}\<[E]%
\\
\>[B]{}\mathbf{type}\;\Conid{Slot}{}\<[14]%
\>[14]{}\mathrel{=}(\Conid{SID},[\mskip1.5mu \Conid{SID}\mskip1.5mu],[\mskip1.5mu \Conid{SID}\mskip1.5mu]){}\<[E]%
\\
\>[B]{}\mathbf{type}\;\Conid{Comm}{}\<[14]%
\>[14]{}\mathrel{=}(\Conid{SID},\Conid{Int}){}\<[29]%
\>[29]{}\mbox{\onelinecomment  commencement}{}\<[E]%
\\
\>[B]{}\mathbf{type}\;\Conid{CID}{}\<[14]%
\>[14]{}\mathrel{=}(\Conid{Slot},\Conid{Int}){}\<[29]%
\>[29]{}\mbox{\onelinecomment  continuation identifier}{}\<[E]%
\\
\>[B]{}\mathbf{type}\;\Conid{CMap}\;\Varid{t}{}\<[14]%
\>[14]{}\mathrel{=}\Conid{Map}\;\Conid{CID}\;(\Conid{Cont}\;\Varid{t}){}\<[E]%
\\
\>[B]{}\mathbf{type}\;\Conid{USet}{}\<[14]%
\>[14]{}\equiv \Conid{Set}\;\Conid{Descr}{}\<[E]%
\\
\>[B]{}\Varid{addDescr}{}\<[14]%
\>[14]{}\mathbin{::}\Conid{Descr}{}\<[24]%
\>[24]{}\to \Conid{USet}{}\<[34]%
\>[34]{}\to \Conid{USet}{}\<[E]%
\\
\>[B]{}\Varid{hasDescr}{}\<[14]%
\>[14]{}\mathbin{::}\Conid{Descr}{}\<[24]%
\>[24]{}\to \Conid{USet}{}\<[34]%
\>[34]{}\to \Conid{Bool}{}\<[E]%
\\
\>[B]{}\mathbf{type}\;\Conid{GRel}{}\<[14]%
\>[14]{}\equiv \Conid{Set}\;(\Conid{Comm},\Conid{CID}){}\<[E]%
\\
\>[B]{}\Varid{addCont}{}\<[14]%
\>[14]{}\mathbin{::}\Conid{Comm}{}\<[23]%
\>[23]{}\to \Conid{CID}\to \Conid{GRel}\to \Conid{GRel}{}\<[E]%
\\
\>[B]{}\Varid{conts}{}\<[14]%
\>[14]{}\mathbin{::}\Conid{Comm}{}\<[23]%
\>[23]{}\to \Conid{GRel}{}\<[32]%
\>[32]{}\to [\mskip1.5mu \Conid{CID}\mskip1.5mu]{}\<[E]%
\\
\>[B]{}\mathbf{type}\;\Conid{PRel}{}\<[14]%
\>[14]{}\equiv \Conid{Set}\;(\Conid{Comm},\Conid{Int}){}\<[E]%
\\
\>[B]{}\Varid{addExtent}{}\<[14]%
\>[14]{}\mathbin{::}\Conid{Comm}{}\<[23]%
\>[23]{}\to \Conid{Int}\to \Conid{PRel}\to \Conid{PRel}{}\<[E]%
\\
\>[B]{}\Varid{extents}{}\<[14]%
\>[14]{}\mathbin{::}\Conid{Comm}{}\<[23]%
\>[23]{}\to \Conid{PRel}{}\<[32]%
\>[32]{}\to [\mskip1.5mu \Conid{Int}\mskip1.5mu]{}\<[E]%
\\
\>[B]{}\mathbf{type}\;\Conid{BSRs}{}\<[14]%
\>[14]{}\equiv \Conid{Set}\;\Conid{BSR}{}\<[E]%
\\
\>[B]{}\mathbf{type}\;\Conid{BSR}{}\<[14]%
\>[14]{}\mathrel{=}{}\<[17]%
\>[17]{}(\Conid{Slot},\Conid{Int},\Conid{Int},\Conid{Int}){}\<[E]%
\\
\>[B]{}\Varid{addBSR}{}\<[14]%
\>[14]{}\mathbin{::}\Conid{BSR}\to \Conid{BSRs}\to \Conid{BSRs}{}\<[E]%
\\
\>[B]{}\Varid{pivots}{}\<[14]%
\>[14]{}\mathbin{::}(\Conid{Slot},\Conid{Int},\Conid{Int})\to \Conid{BSRs}\to [\mskip1.5mu \Conid{Int}\mskip1.5mu]{}\<[E]%
\ColumnHook
\end{hscode}\resethooks
\caption{The \fungll{} datastructures and operations.}%
\label{fig_fungll_datas}%
\end{figure}%
\begin{figure}[t]%
\begin{hscode}\SaveRestoreHook
\column{B}{@{}>{\hspre}l<{\hspost}@{}}%
\column{9}{@{}>{\hspre}c<{\hspost}@{}}%
\column{9E}{@{}l@{}}%
\column{13}{@{}>{\hspre}l<{\hspost}@{}}%
\column{32}{@{}>{\hspre}l<{\hspost}@{}}%
\column{E}{@{}>{\hspre}l<{\hspost}@{}}%
\>[B]{}\Conid{CSV}\;(\Varid{v}){}\<[9]%
\>[9]{}\mathbin{:}{}\<[9E]%
\>[13]{}\Conid{CSV}\;(\Varid{v})\;\text{\tt ','}\;\Conid{CSV}\;(\Varid{v})\;{}\<[32]%
\>[32]{}\{\mskip1.5mu \mathbin{\$}\mathrm{1}\plus \mathbin{\$}\mathrm{3}\mskip1.5mu\}{}\<[E]%
\\
\>[9]{}\mid {}\<[9E]%
\>[13]{}\Varid{v}\;{}\<[32]%
\>[32]{}\{\mskip1.5mu [\mskip1.5mu \mathbin{\$}\mathrm{1}\mskip1.5mu]\mskip1.5mu\}{}\<[E]%
\ColumnHook
\end{hscode}\resethooks
\caption{A left-recursive parameterized nonterminal.}%
\label{fig_csv}%
\end{figure}%
\begin{figure}[t]%
\begin{hscode}\SaveRestoreHook
\footnotesize{}
\column{B}{@{}>{\hspre}l<{\hspost}@{}}%
\column{3}{@{}>{\hspre}l<{\hspost}@{}}%
\column{5}{@{}>{\hspre}l<{\hspost}@{}}%
\column{6}{@{}>{\hspre}l<{\hspost}@{}}%
\column{7}{@{}>{\hspre}l<{\hspost}@{}}%
\column{14}{@{}>{\hspre}c<{\hspost}@{}}%
\column{14E}{@{}l@{}}%
\column{18}{@{}>{\hspre}l<{\hspost}@{}}%
\column{24}{@{}>{\hspre}l<{\hspost}@{}}%
\column{34}{@{}>{\hspre}l<{\hspost}@{}}%
\column{38}{@{}>{\hspre}l<{\hspost}@{}}%
\column{54}{@{}>{\hspre}l<{\hspost}@{}}%
\column{E}{@{}>{\hspre}l<{\hspost}@{}}%
\>[B]{}\Varid{sCSV}\;\Varid{x\char95 1}\mathrel{=}\Conid{Symbol}\;\Varid{nt}\;\Varid{matcher}\;\Varid{evaluator}{}\<[E]%
\\
\>[B]{}\hsindent{3}{}\<[3]%
\>[3]{}\mathbf{where}{}\<[E]%
\\
\>[3]{}\hsindent{2}{}\<[5]%
\>[5]{}\Varid{nt}\mathrel{=}\Conid{App}\;\text{\tt \char34 CSV\char34}\;[\mskip1.5mu \Varid{id}_\mathit{sm}\;\Varid{x\char95 1}\mskip1.5mu]{}\<[E]%
\\
\>[3]{}\hsindent{2}{}\<[5]%
\>[5]{}\Varid{matcher}\;\Varid{l}\;(\Varid{c},\Varid{cf})\mathrel{=}\Varid{descend}\;(\Varid{nt},\Varid{l})\;(\Varid{c},\Varid{cf})\;(c_0\;\Varid{l}\;\Varid{l}\mathbin{\circ}\Varid{c\char95 4}\;\Varid{l}\;\Varid{l}){}\<[E]%
\\
\>[5]{}\hsindent{1}{}\<[6]%
\>[6]{}\mathbf{where}{}\<[E]%
\\
\>[6]{}\hsindent{1}{}\<[7]%
\>[7]{}c_0\;\Varid{k}\;\Varid{r}\mathrel{=}\Varid{continue}\;(\Varid{s\char95 0},\Varid{l},\Varid{k},\Varid{r})\;(\Varid{match}_\mathit{sm}\;(\Varid{sCSV}\;\Varid{x\char95 1})\;\Varid{r}\;((\Varid{s\char95 1},\Varid{l}),c_1)){}\<[E]%
\\
\>[6]{}\hsindent{1}{}\<[7]%
\>[7]{}c_1\;\Varid{k}\;\Varid{r}\mathrel{=}\Varid{continue}\;(\Varid{s\char95 1},\Varid{l},\Varid{k},\Varid{r})\;(\Varid{match}_\mathit{sm}\;\Varid{sCom}\;\Varid{r}\;((\Varid{s\char95 2},\Varid{l}),c_2)){}\<[E]%
\\
\>[6]{}\hsindent{1}{}\<[7]%
\>[7]{}c_2\;\Varid{k}\;\Varid{r}\mathrel{=}\Varid{continue}\;(\Varid{s\char95 2},\Varid{l},\Varid{k},\Varid{r})\;(\Varid{match}_\mathit{sm}\;(\Varid{sCSV}\;\Varid{x\char95 1})\;\Varid{r}\;((\Varid{s\char95 3},\Varid{l}),\Varid{c\char95 3})){}\<[E]%
\\
\>[6]{}\hsindent{1}{}\<[7]%
\>[7]{}\Varid{c\char95 3}\;\Varid{k}\;\Varid{r}\mathrel{=}\Varid{continue}\;(\Varid{s\char95 3},\Varid{l},\Varid{k},\Varid{r})\;(\Varid{ascend}\;(\Varid{nt},\Varid{l})\;\Varid{r}){}\<[E]%
\\
\>[6]{}\hsindent{1}{}\<[7]%
\>[7]{}\Varid{c\char95 4}\;\Varid{k}\;\Varid{r}\mathrel{=}\Varid{continue}\;(\Varid{s\char95 4},\Varid{l},\Varid{k},\Varid{r})\;(\Varid{match}_\mathit{sm}\;\Varid{x\char95 1}\;\Varid{r}\;((\Varid{s\char95 5},\Varid{l}),\Varid{c\char95 5})){}\<[E]%
\\
\>[6]{}\hsindent{1}{}\<[7]%
\>[7]{}\Varid{c\char95 5}\;\Varid{k}\;\Varid{r}\mathrel{=}\Varid{continue}\;(\Varid{s\char95 5},\Varid{l},\Varid{k}\;\Varid{r})\;(\Varid{ascend}\;(\Varid{nt},\Varid{l})\;\Varid{r}){}\<[E]%
\\
\>[3]{}\hsindent{2}{}\<[5]%
\>[5]{}\Varid{s\char95 0}\mathrel{=}(\Varid{nt},[\mskip1.5mu \mskip1.5mu],[\mskip1.5mu \Varid{id}_\mathit{sm}\;(\Varid{sCSV}\;\Varid{x\char95 1}),\Varid{id}_\mathit{sm}\;\Varid{sCom},\Varid{id}_\mathit{sm}\;(\Varid{sCSV}\;\Varid{x\char95 1})\mskip1.5mu]){}\<[E]%
\\
\>[3]{}\hsindent{2}{}\<[5]%
\>[5]{}\Varid{s\char95 1}\mathrel{=}(\Varid{nt},[\mskip1.5mu \Varid{id}_\mathit{sm}\;(\Varid{sCSV}\;\Varid{x\char95 1})\mskip1.5mu],[\mskip1.5mu \Varid{id}_\mathit{sm}\;\Varid{sCom},\Varid{id}_\mathit{sm}\;(\Varid{sCSV}\;\Varid{x\char95 1})\mskip1.5mu]){}\<[E]%
\\
\>[3]{}\hsindent{2}{}\<[5]%
\>[5]{}\Varid{s\char95 2}\mathrel{=}(\Varid{nt},[\mskip1.5mu \Varid{id}_\mathit{sm}\;(\Varid{sCSV}\;\Varid{x\char95 1}),\Varid{id}_\mathit{sm}\;\Varid{sCom}\mskip1.5mu],[\mskip1.5mu \Varid{id}_\mathit{sm}\;(\Varid{sCSV}\;\Varid{x\char95 1})\mskip1.5mu]){}\<[E]%
\\
\>[3]{}\hsindent{2}{}\<[5]%
\>[5]{}\Varid{s\char95 3}\mathrel{=}(\Varid{nt},[\mskip1.5mu \Varid{id}_\mathit{sm}\;(\Varid{sCSV}\;\Varid{x\char95 1}),\Varid{id}_\mathit{sm}\;\Varid{sCom},\Varid{id}_\mathit{sm}\;(\Varid{sCSV}\;\Varid{x\char95 1})\mskip1.5mu],[\mskip1.5mu \mskip1.5mu]){}\<[E]%
\\
\>[3]{}\hsindent{2}{}\<[5]%
\>[5]{}\Varid{s\char95 4}\mathrel{=}(\Varid{nt},[\mskip1.5mu \mskip1.5mu],[\mskip1.5mu \Varid{id}_\mathit{sm}\;\Varid{x\char95 1}\mskip1.5mu]){}\<[E]%
\\
\>[3]{}\hsindent{2}{}\<[5]%
\>[5]{}\Varid{s\char95 5}\mathrel{=}(\Varid{nt},[\mskip1.5mu \Varid{id}_\mathit{sm}\;\Varid{x\char95 1}\mskip1.5mu],[\mskip1.5mu \mskip1.5mu]){}\<[E]%
\\
\>[3]{}\hsindent{2}{}\<[5]%
\>[5]{}\Varid{evaluator}\;\Varid{inp}\;\Varid{bsrs}\;\Varid{l}\;\Varid{r}\;\Varid{nts}{}\<[34]%
\>[34]{}\mid \Varid{nt}\in\Varid{nts}{}\<[54]%
\>[54]{}\mathrel{=}[\mskip1.5mu \mskip1.5mu]{}\<[E]%
\\
\>[34]{}\mid \Varid{otherwise}{}\<[54]%
\>[54]{}\mathrel{=}\Varid{res}{}\<[E]%
\\
\>[5]{}\hsindent{1}{}\<[6]%
\>[6]{}\mathbf{where}{}\<[E]%
\\
\>[6]{}\hsindent{1}{}\<[7]%
\>[7]{}\Varid{nts'}\;\Varid{p}\;\Varid{q}{}\<[18]%
\>[18]{}\mid \Varid{l}\not\equiv \Varid{p}\mathrel{\vee}\Varid{r}\not\equiv \Varid{q}{}\<[38]%
\>[38]{}\mathrel{=}\Varid{empty}{}\<[E]%
\\
\>[18]{}\mid \Varid{otherwise}{}\<[38]%
\>[38]{}\mathrel{=}\Varid{insert}\;\Varid{nt}\;\Varid{nts}{}\<[E]%
\\
\>[6]{}\hsindent{1}{}\<[7]%
\>[7]{}\Varid{res}\mathrel{=}{}\<[14]%
\>[14]{}[\mskip1.5mu {}\<[14E]%
\>[18]{}\Varid{sv\char95 1}\plus \Varid{sv\char95 3}\mbox{\onelinecomment  semantic action of first alternate}{}\<[E]%
\\
\>[14]{}\mid {}\<[14E]%
\>[18]{}\Varid{p\char95 3}{}\<[24]%
\>[24]{}\leftarrow \Varid{pivots}\;(\Varid{s\char95 3},\Varid{l},\Varid{r})\;\Varid{bsrs}{}\<[E]%
\\
\>[14]{},{}\<[14E]%
\>[18]{}\Varid{p\char95 2}{}\<[24]%
\>[24]{}\leftarrow \Varid{pivots}\;(\Varid{s\char95 2},\Varid{l},\Varid{p\char95 3})\;\Varid{bsrs}{}\<[E]%
\\
\>[14]{},{}\<[14E]%
\>[18]{}\Varid{p\char95 1}{}\<[24]%
\>[24]{}\leftarrow \Varid{pivots}\;(\Varid{s\char95 1},\Varid{l},\Varid{p\char95 2})\;\Varid{bsrs}\mbox{\onelinecomment  \ensuremath{\Varid{p\char95 1}\equiv \Varid{l}}}{}\<[E]%
\\
\>[14]{},{}\<[14E]%
\>[18]{}\Varid{sv\char95 3}{}\<[24]%
\>[24]{}\leftarrow \Varid{eval}_\mathit{sm}\;(\Varid{sCSV}\;\Varid{x\char95 1})\;\Varid{inp}\;\Varid{bsrs}\;\Varid{p\char95 3}\;\Varid{r}\;(\Varid{nts'}\;\Varid{p\char95 3}\;\Varid{r}){}\<[E]%
\\
\>[14]{},{}\<[14E]%
\>[18]{}\Varid{sv\char95 2}{}\<[24]%
\>[24]{}\leftarrow \Varid{eval}_\mathit{sm}\;\Varid{sCom}\;\Varid{inp}\;\Varid{bsrs}\;\Varid{p\char95 2}\;\Varid{p\char95 3}\;(\Varid{nts'}\;\Varid{p\char95 2}\;\Varid{p\char95 3}){}\<[E]%
\\
\>[14]{},{}\<[14E]%
\>[18]{}\Varid{sv\char95 1}{}\<[24]%
\>[24]{}\leftarrow \Varid{eval}_\mathit{sm}\;(\Varid{sCSV}\;\Varid{x\char95 1})\;\Varid{inp}\;\Varid{bsrs}\;\Varid{l}\;\Varid{p\char95 2}\;(\Varid{nts'}\;\Varid{l}\;\Varid{p\char95 2}){}\<[E]%
\\
\>[14]{}\mskip1.5mu]{}\<[14E]%
\>[18]{}\plus {}\<[E]%
\\
\>[14]{}[\mskip1.5mu {}\<[14E]%
\>[18]{}[\mskip1.5mu \Varid{sv\char95 1}\mskip1.5mu]\mbox{\onelinecomment  semantic action of second alternate}{}\<[E]%
\\
\>[14]{},{}\<[14E]%
\>[18]{}\Varid{p\char95 1}{}\<[24]%
\>[24]{}\leftarrow \Varid{pivots}\;(\Varid{s\char95 5},\Varid{l},\Varid{r})\;\Varid{bsrs}\mbox{\onelinecomment  \ensuremath{\Varid{p\char95 1}\equiv \Varid{l}}}{}\<[E]%
\\
\>[14]{}\mid {}\<[14E]%
\>[18]{}\Varid{sv\char95 1}{}\<[24]%
\>[24]{}\leftarrow \Varid{eval}_\mathit{sm}\;\Varid{x\char95 1}\;\Varid{inp}\;\Varid{bsrs}\;\Varid{l}\;\Varid{r}\;(\Varid{nts'}\;\Varid{l}\;\Varid{r})\mskip1.5mu]{}\<[E]%
\ColumnHook
\end{hscode}\resethooks
\caption{Code generated by the GLL back-end for the production of Figure~\ref{fig_csv}. \ensuremath{\Varid{sCom}} is the \ensuremath{\Conid{Symbol}} for token \ensuremath{\text{\tt ','}}.}%
\label{fig_csv_code}%
\end{figure}%
This subsection explains, by example, how the GLL back-end translates the parameterized nonterminals of Happy grammars to higher-order functions that implement the \fungll{} algorithm and explains how the aforementioned `semantic phase' is implemented.
The example is provided by the definition of \ensuremath{\Conid{CSV}} in Figure~\ref{fig_csv} and the generated code in Figure~\ref{fig_csv_code}.
The translation of \ensuremath{\%\textbf{token}} directives (and other Happy directives) has been omitted.
Figure~\ref{fig_fungll_datas} shows the data\-structures\footnote{Some type definitions use \ensuremath{\equiv } rather than \ensuremath{\mathrel{=}} to indicate that the datastructures are equivalent to sets but are implemented differently for efficiently.} and operations of the \fungll{} algorithm as provided by the support library of the implementation.
Every \ensuremath{\%\textbf{token}} directive and every nonterminal definition of a Happy grammar generates a function that returns a \ensuremath{\Conid{Symbol}}.
If generated for a parameterized nonterminal, the function has a parameter of type \ensuremath{\Conid{Symbol}} for every parameter of the nonterminal.
As the example shows, when a symbol is used in the code generated for another symbol, it is irrelevant whether the symbol is made available as a parameter or is defined in the same namespace.
\begin{hscode}\SaveRestoreHook
\column{B}{@{}>{\hspre}l<{\hspost}@{}}%
\column{7}{@{}>{\hspre}l<{\hspost}@{}}%
\column{9}{@{}>{\hspre}c<{\hspost}@{}}%
\column{9E}{@{}l@{}}%
\column{13}{@{}>{\hspre}l<{\hspost}@{}}%
\column{19}{@{}>{\hspre}c<{\hspost}@{}}%
\column{19E}{@{}l@{}}%
\column{22}{@{}>{\hspre}l<{\hspost}@{}}%
\column{24}{@{}>{\hspre}l<{\hspost}@{}}%
\column{35}{@{}>{\hspre}l<{\hspost}@{}}%
\column{46}{@{}>{\hspre}l<{\hspost}@{}}%
\column{62}{@{}>{\hspre}l<{\hspost}@{}}%
\column{73}{@{}>{\hspre}l<{\hspost}@{}}%
\column{E}{@{}>{\hspre}l<{\hspost}@{}}%
\>[B]{}\mathbf{data}\;{}\<[7]%
\>[7]{}\Conid{Symbol}\;\Varid{t}\;\Varid{a}{}\<[19]%
\>[19]{}\mathrel{=}{}\<[19E]%
\>[22]{}\Conid{Symbol}{}\<[E]%
\\
\>[7]{}\hsindent{2}{}\<[9]%
\>[9]{}\{\mskip1.5mu {}\<[9E]%
\>[13]{}\Varid{id}_\mathit{sm}{}\<[24]%
\>[24]{}\mathbin{::}\Conid{SID},{}\<[35]%
\>[35]{}\Varid{match}_\mathit{sm}{}\<[46]%
\>[46]{}\mathbin{::}\Conid{Matcher}\;\Varid{t},{}\<[62]%
\>[62]{}\Varid{eval}_\mathit{sm}{}\<[73]%
\>[73]{}\mathbin{::}\Conid{Evaluator}\;\Varid{t}\;\Varid{a}\mskip1.5mu\}{}\<[E]%
\ColumnHook
\end{hscode}\resethooks
Besides an identifier, every \ensuremath{\Conid{Symbol}} consists of a `matcher' and an `evaluator', implementing \fungll{} and the semantic phase respectively.
A symbol identifier (\ensuremath{\Conid{SID}}, see Figure~\ref{fig_fungll_datas}) is either a token name (\ensuremath{\Conid{Tn}}) or the name of a nonterminal applied to zero or more symbol identifier arguments (\ensuremath{\Conid{App}}).
%
%
%
The type parameters \ensuremath{\Varid{t}} and \ensuremath{\Varid{a}} of \ensuremath{\Conid{Symbol}} are for the type of tokens and the type of semantic values produced by the evaluator.
The generated code in Figure~\ref{fig_csv_code} shows how the slots (named $s_0,...,s_5$ in the code) for the definition of \ensuremath{\Conid{CSV}} are computed based on the symbols and the parameters that are mentioned in the definition of \ensuremath{\Conid{CSV}}.

A matcher (\ensuremath{\Conid{Matcher}}) receives an index (pivot or right extent) and a continuation paired with its continuation identifier (of type \ensuremath{\Conid{CID}}).
A matcher exhibits the necessary effects on the \fungll{} datastructures (\ensuremath{\Conid{Data}}) as prescribed by either the \textbf{continue}, \textbf{ascend} or \textbf{descend} action, explained in the previous subsection.
An immutable array holds the input sentence for fast access.
\begin{hscode}\SaveRestoreHook
\column{B}{@{}>{\hspre}l<{\hspost}@{}}%
\column{8}{@{}>{\hspre}l<{\hspost}@{}}%
\column{19}{@{}>{\hspre}l<{\hspost}@{}}%
\column{23}{@{}>{\hspre}c<{\hspost}@{}}%
\column{23E}{@{}l@{}}%
\column{26}{@{}>{\hspre}l<{\hspost}@{}}%
\column{27}{@{}>{\hspre}l<{\hspost}@{}}%
\column{32}{@{}>{\hspre}l<{\hspost}@{}}%
\column{33}{@{}>{\hspre}l<{\hspost}@{}}%
\column{44}{@{}>{\hspre}l<{\hspost}@{}}%
\column{45}{@{}>{\hspre}l<{\hspost}@{}}%
\column{50}{@{}>{\hspre}l<{\hspost}@{}}%
\column{51}{@{}>{\hspre}l<{\hspost}@{}}%
\column{60}{@{}>{\hspre}l<{\hspost}@{}}%
\column{62}{@{}>{\hspre}l<{\hspost}@{}}%
\column{67}{@{}>{\hspre}l<{\hspost}@{}}%
\column{68}{@{}>{\hspre}l<{\hspost}@{}}%
\column{E}{@{}>{\hspre}l<{\hspost}@{}}%
\>[B]{}\mathbf{type}\;{}\<[8]%
\>[8]{}\Conid{Matcher}\;\Varid{t}{}\<[19]%
\>[19]{}\mathrel{=}\Conid{Int}{}\<[27]%
\>[27]{}\to (\Conid{CID},\Conid{Cont}\;\Varid{t}){}\<[45]%
\>[45]{}\to \Conid{Data}\;\Varid{t}\to \Conid{Data}\;\Varid{t}{}\<[E]%
\\
\>[B]{}\mathbf{type}\;{}\<[8]%
\>[8]{}\Conid{Cont}\;\Varid{t}{}\<[19]%
\>[19]{}\mathrel{=}\Conid{Int}{}\<[27]%
\>[27]{}\to \Conid{Int}{}\<[45]%
\>[45]{}\to \Conid{Data}\;\Varid{t}\to \Conid{Data}\;\Varid{t}{}\<[E]%
\\
\>[B]{}\mathbf{data}\;{}\<[8]%
\>[8]{}\Conid{Data}\;\Varid{t}\mathrel{=}\Conid{Data}\;{}\<[23]%
\>[23]{}\{\mskip1.5mu {}\<[23E]%
\>[26]{}\Varid{uset}{}\<[33]%
\>[33]{}\mathbin{::}\Conid{USet},{}\<[44]%
\>[44]{}\Varid{bsrs}{}\<[50]%
\>[50]{}\mathbin{::}\Conid{BSRs},{}\<[60]%
\>[60]{}\Varid{grel}{}\<[67]%
\>[67]{}\mathbin{::}\Conid{GRel}{}\<[E]%
\\
\>[23]{},{}\<[23E]%
\>[26]{}\Varid{cmap}{}\<[32]%
\>[32]{}\mathbin{::}\Conid{CMap}\;\Varid{t},{}\<[44]%
\>[44]{}\Varid{prel}{}\<[51]%
\>[51]{}\mathbin{::}\Conid{PRel},{}\<[62]%
\>[62]{}\Varid{inp}{}\<[68]%
\>[68]{}\mathbin{::}\Conid{Array}\;\Conid{Int}\;\Varid{t}\mskip1.5mu\}{}\<[E]%
\ColumnHook
\end{hscode}\resethooks
A continuation identifier is a pair of a slot and a left extent, as in the previous subsection.
The continuation itself (\ensuremath{\Conid{Cont}}) captures the behavior of the \textbf{continue} action of the previous section.
The functions $c_0,\ldots,c_5$ in the generated code are the continuation functions defined for \ensuremath{\Conid{CSV}}, with one for every slot.
These definitions are local to the definition of the matcher for \ensuremath{\Conid{CSV}} because the matcher is called with the left extent $l$.
The slot for which the continuation is defined (e.g. $s_0$ for $c_0$), the left extent $l$ (parameter of \ensuremath{\Varid{matcher}}) and the parameters of the continuation (\ensuremath{\Varid{k}} and \ensuremath{\Varid{r}}) form the descriptor and BSR element required to perform the \textbf{continue} action.
The support function \ensuremath{\Varid{continue}} receives both as a single (BSR) argument and implements the behavior of \textbf{continue}.
\begin{hscode}\SaveRestoreHook
\column{B}{@{}>{\hspre}l<{\hspost}@{}}%
\column{3}{@{}>{\hspre}c<{\hspost}@{}}%
\column{3E}{@{}l@{}}%
\column{6}{@{}>{\hspre}l<{\hspost}@{}}%
\column{34}{@{}>{\hspre}l<{\hspost}@{}}%
\column{E}{@{}>{\hspre}l<{\hspost}@{}}%
\>[B]{}\Varid{continue}\mathbin{::}\Conid{BSR}\to (\Conid{Data}\;\Varid{t}\to \Conid{Data}\;\Varid{t})\to \Conid{Data}\;\Varid{t}\to \Conid{Data}\;\Varid{t}{}\<[E]%
\\
\>[B]{}\Varid{continue}\;\Varid{bsr}\;\Varid{cf}\;\Varid{d}\mathrel{=}\Varid{maybeProcess}\;\Varid{bsr}\;\Varid{cf}\;(\Varid{addBSR}\;\Varid{bsr}\;\Varid{d}){}\<[E]%
\\[\blanklineskip]%
\>[B]{}\Varid{maybeProcess}\;(\Varid{s},\Varid{l},\Varid{k},\Varid{r})\;\Varid{cf}\;\Varid{d}{}\<[E]%
\\
\>[B]{}\hsindent{3}{}\<[3]%
\>[3]{}\mid {}\<[3E]%
\>[6]{}\Varid{hasDescr}\;(\Varid{s},\Varid{l},\Varid{r})\;(\Varid{uset}\;\Varid{d}){}\<[34]%
\>[34]{}\mathrel{=}\Varid{d}{}\<[E]%
\\
\>[B]{}\hsindent{3}{}\<[3]%
\>[3]{}\mid {}\<[3E]%
\>[6]{}\Varid{otherwise}\mathrel{=}\Varid{cf}\;(\Varid{d}\;\{\mskip1.5mu \Varid{uset}\mathrel{=}\Varid{addDescr}\;(\Varid{s},\Varid{l},\Varid{r})\;(\Varid{uset}\;\Varid{d})\mskip1.5mu\}){}\<[E]%
\ColumnHook
\end{hscode}\resethooks
The BSR element is always added to \ensuremath{\Varid{bsrs}} by \ensuremath{\Varid{continue}}, whereas \ensuremath{\Varid{cf}} (capturing the effect of the next processing step) is only applied if the descriptor is not already in \ensuremath{\Varid{uset}}.
The second argument of \ensuremath{\Varid{continue}} is the result (a function \ensuremath{\Conid{Data}\;\Varid{t}\to \Conid{Data}\;\Varid{t}}) of applying the matcher for the next symbol to be matched according to the slot, i.e. the first symbol in the third component of the slot (for example, the parameter $x_1$ in the case of $c_4$ and $s_4$).
The matcher is applied to the right extent $r$ and the continuation and continuation identifier pair for the `next' slot (e.g. $s_1$ in the case of $s_0$, $s_2$ in the case of $s_1$, etc.) with the same left extent. 
Or, if the third component of the slot is empty (e.g. in the case of $s_3$ and $s_5$), the support function \ensuremath{\Varid{ascend}} is applied, corresponding to the \textbf{ascend} action.
In other words, the continuation function generated for slot \ensuremath{(\Varid{x},\alpha,\beta)} supplies the matcher function generated for the first symbol of \ensuremath{\beta} to \ensuremath{\Varid{continue}} or supplies the function \ensuremath{\Varid{ascend}} if there is no such symbol.
In this way, the \textbf{process} action is performed.

The function \ensuremath{\Varid{ascend}} is given a commencement \ensuremath{(\Varid{nt},\Varid{l})} and a right extent \ensuremath{\Varid{r}} and performs the \textbf{ascend} action.
The \textbf{ascend} action involves looking up all continuation identifiers $c$ stored in \ensuremath{\Varid{grel}} for the given commencement.
The continuation identified by \ensuremath{\Varid{c}} is recorded in the map \ensuremath{\Varid{cmap}} (see the definition of \ensuremath{\Varid{descend}} below).
\begin{hscode}\SaveRestoreHook
\column{B}{@{}>{\hspre}l<{\hspost}@{}}%
\column{3}{@{}>{\hspre}l<{\hspost}@{}}%
\column{11}{@{}>{\hspre}l<{\hspost}@{}}%
\column{16}{@{}>{\hspre}l<{\hspost}@{}}%
\column{E}{@{}>{\hspre}l<{\hspost}@{}}%
\>[B]{}\Varid{ascend}\mathbin{::}\Conid{Comm}\to \Conid{Int}\to \Conid{Data}\;\Varid{t}\to \Conid{Data}\;\Varid{t}{}\<[E]%
\\
\>[B]{}\Varid{ascend}\;(\Varid{nt},\Varid{l})\;\Varid{r}\;\Varid{d}\mathrel{=}\Varid{foldr}\;((\mathbin{\$})\mathbin{\circ}(\lambda \Varid{cf}\to \Varid{cf}\;\Varid{l}\;\Varid{r}))\;\Varid{id}\;\Varid{cs}\;\Varid{d'}{}\<[E]%
\\
\>[B]{}\hsindent{3}{}\<[3]%
\>[3]{}\mathbf{where}\;{}\<[11]%
\>[11]{}\Varid{d'}{}\<[16]%
\>[16]{}\mathrel{=}\Varid{d}\;\{\mskip1.5mu \Varid{prel}\mathrel{=}\Varid{addExtent}\;(\Varid{nt},\Varid{l})\;\Varid{r}\;(\Varid{prel}\;\Varid{d})\mskip1.5mu\}{}\<[E]%
\\
\>[11]{}\Varid{cs}{}\<[16]%
\>[16]{}\mathrel{=}\Varid{map}\;(\Varid{flip}\;\Varid{lookup}\;(\Varid{cmap}\;\Varid{d}))\;(\Varid{conts}\;(\Varid{nt},\Varid{l})\;(\Varid{grel}\;\Varid{d})){}\<[E]%
\ColumnHook
\end{hscode}\resethooks
The effects of the different continuations applied to $l$ and $r$ (functions from \ensuremath{\Conid{Data}\to \Conid{Data}}, one for every \ensuremath{\Varid{c}}) accumulate by function composition.
The order of composition does not influence the outcome since descriptors can be processed in any order by \fungll{}.

The matcher for a token (not shown here) is defined in terms of a support function like \ensuremath{\Varid{matchPattern}} (see Section~\ref{sec_motivation}).
The matcher for a nonterminal with name \ensuremath{\Varid{nt}} applies \ensuremath{\Varid{descend}} (see the definition of \ensuremath{\Varid{matcher}} for \ensuremath{\Conid{CSV}}) to commencement \ensuremath{(\Varid{nt},\Varid{l})} and a continuation \ensuremath{\Varid{cf}} paired with its identifier \ensuremath{\Varid{c}}.
The arguments \ensuremath{\Varid{l}} and \ensuremath{(\Varid{c},\Varid{cf})} are themselves inputs of the matcher, supplied when continuations are applied (see the definitions of $c_0,\ldots,c_5$ in the code generated for \ensuremath{\Conid{CSV}}).
Function \ensuremath{\Varid{descend}} also receives the effects of processing the alternates of \ensuremath{\Varid{nt}}.
In the case of the example, the effects of the alternates of \ensuremath{\Conid{CSV}} are produced by composing the effects of applying the continuations $c_0$ and $c_4$ that match the first symbol of each of the two alternates.
\begin{hscode}\SaveRestoreHook
\column{B}{@{}>{\hspre}l<{\hspost}@{}}%
\column{3}{@{}>{\hspre}l<{\hspost}@{}}%
\column{11}{@{}>{\hspre}l<{\hspost}@{}}%
\column{16}{@{}>{\hspre}l<{\hspost}@{}}%
\column{17}{@{}>{\hspre}l<{\hspost}@{}}%
\column{19}{@{}>{\hspre}l<{\hspost}@{}}%
\column{23}{@{}>{\hspre}l<{\hspost}@{}}%
\column{31}{@{}>{\hspre}l<{\hspost}@{}}%
\column{E}{@{}>{\hspre}l<{\hspost}@{}}%
\>[B]{}\Varid{descend}\mathbin{::}\Conid{Comm}{}\<[19]%
\>[19]{}\to (\Conid{CID},\Conid{Cont}\;\Varid{t})\to (\Conid{Data}\;\Varid{t}\to \Conid{Data}\;\Varid{t}){}\<[E]%
\\
\>[19]{}\to \Conid{Data}\;\Varid{t}\to \Conid{Data}\;\Varid{t}{}\<[E]%
\\
\>[B]{}\Varid{descend}\;(\Varid{x},\Varid{l})\;(\Varid{c},\Varid{cf})\;\Varid{alts}\;\Varid{d}{}\<[E]%
\\
\>[B]{}\mid \Varid{null}\;\Varid{rs}{}\<[16]%
\>[16]{}\mathrel{=}\Varid{alts}\;\Varid{d'}{}\<[E]%
\\
\>[B]{}\mid \Varid{otherwise}{}\<[16]%
\>[16]{}\mathrel{=}\Varid{foldr}\;((\mathbin{\$})\mathbin{\circ}(\lambda \Varid{r}\to \Varid{cf}\;\Varid{l}\;\Varid{r}))\;\Varid{id}\;\Varid{rs}\;\Varid{d'}{}\<[E]%
\\
\>[B]{}\hsindent{3}{}\<[3]%
\>[3]{}\mathbf{where}\;{}\<[11]%
\>[11]{}\Varid{rs}{}\<[17]%
\>[17]{}\mathrel{=}\Varid{extents}\;(\Varid{x},\Varid{l})\;(\Varid{prel}\;\Varid{d}){}\<[E]%
\\
\>[11]{}\Varid{d'}{}\<[17]%
\>[17]{}\mathrel{=}\Varid{d}\;{}\<[23]%
\>[23]{}\{\mskip1.5mu \Varid{grel}{}\<[31]%
\>[31]{}\mathrel{=}\Varid{addCont}\;(\Varid{x},\Varid{l})\;\Varid{c}\;(\Varid{grel}\;\Varid{d}){}\<[E]%
\\
\>[23]{},\Varid{cmap}{}\<[31]%
\>[31]{}\mathrel{=}\Varid{insert}\;\Varid{c}\;\Varid{cf}\;(\Varid{cmap}\;\Varid{d})\mskip1.5mu\}{}\<[E]%
\ColumnHook
\end{hscode}\resethooks
Depending on whether there are right extents (\ensuremath{\Varid{rs}}) stored in \ensuremath{\Varid{prel}} for the given commencement \ensuremath{(\Varid{nt},\Varid{l})}, \ensuremath{\Varid{descend}} either applies the continuation \ensuremath{\Varid{cf}} to each of the right extents or applies the effects of the alternates (\ensuremath{\Varid{alts}}).
In both cases \ensuremath{\Varid{grel}} is extended.

The definition of \ensuremath{\Conid{CSV}} is left-recursive, and its matcher will call itself without consuming tokens from the input sentence (when continuation \ensuremath{c_0} is applied with $r = l$).
This does not result in non-termination however, because the first application of \ensuremath{\Varid{continue}} adds the descriptor \ensuremath{(\Varid{s\char95 0},\Varid{l},\Varid{r})} (with $l = r$) to \ensuremath{\Varid{uset}}, which is `noticed' during the second application.

\paragraph{Semantic phase}
The function \ensuremath{\Varid{run}}, defined\footnote{The support library defines variants of \ensuremath{\Varid{run}} to make partial parsers (and implementing the \ensuremath{\%\textbf{partial}} directive) and parsers that return a value of type \ensuremath{\Conid{Either}\;[\mskip1.5mu \Conid{String}\mskip1.5mu]\;[\mskip1.5mu \Varid{a}\mskip1.5mu]} of which the left component is a sequence of errors.} below, turns a \ensuremath{\Conid{Symbol}} into a parser, a function from a sentence to a list of semantic values.
\begin{hscode}\SaveRestoreHook
\column{B}{@{}>{\hspre}l<{\hspost}@{}}%
\column{3}{@{}>{\hspre}l<{\hspost}@{}}%
\column{11}{@{}>{\hspre}l<{\hspost}@{}}%
\column{E}{@{}>{\hspre}l<{\hspost}@{}}%
\>[B]{}\Varid{run}\mathbin{::}\Conid{Symbol}\;\Varid{t}\;\Varid{a}\to \Conid{Array}\;\Conid{Int}\;\Varid{t}\to [\mskip1.5mu \Varid{a}\mskip1.5mu]{}\<[E]%
\\
\>[B]{}\Varid{run}\;\Varid{x}\;\Varid{str}\mathrel{=}\Varid{eval}_\mathit{sm}\;\Varid{str}\;(\Varid{bsrs}\;\Varid{d\char95 1})\;\mathrm{0}\;(\Varid{length}\;\Varid{str})\;\Varid{empty}{}\<[E]%
\\
\>[B]{}\hsindent{3}{}\<[3]%
\>[3]{}\mathbf{where}\;{}\<[11]%
\>[11]{}\mbox{\onelinecomment  below, all omitted fields are empty}{}\<[E]%
\\
\>[11]{}\Varid{d\char95 0}\mathrel{=}\Conid{Data}\;\{\mskip1.5mu \Varid{inp}\mathrel{=}\Varid{str},\Varid{uset}\mathrel{=}\Varid{empty},\mathbin{...}\mskip1.5mu\}{}\<[E]%
\\
\>[11]{}\mbox{\onelinecomment  below, id is the identity function over \ensuremath{\Conid{Data}\;\Varid{t}}}{}\<[E]%
\\
\>[11]{}\Varid{d\char95 1}\mathrel{=}\Varid{match}_\mathit{sm}\;\mathrm{0}\;(c_0,\Varid{id})\;\Varid{d\char95 0}{}\<[E]%
\\
\>[11]{}\mbox{\onelinecomment  "$\_\_$START" is an artificial start symbol}{}\<[E]%
\\
\>[11]{}c_0\mathrel{=}((\Conid{App}\;\text{\tt \char34 \char95 \char95 START\char34}\;[\mskip1.5mu \mskip1.5mu],[\mskip1.5mu \Varid{id}_\mathit{sm}\;\Varid{x}\mskip1.5mu],[\mskip1.5mu \mskip1.5mu]),\mathrm{0}){}\<[E]%
\ColumnHook
\end{hscode}\resethooks
The result of a parser contains a semantic value for every derivation of the sentence encoded in the BSR set computed by the matcher of the \ensuremath{\Conid{Symbol}}.
The semantic values are based on the semantic actions of the grammar and are computed by the evaluator of the \ensuremath{\Conid{Symbol}}.
Although not shown here, the GLL back-end allows the user to implement disambiguation strategies as filters over semantic values computed from a BSR set.
These strategies can be based on precedence levels, associativity rules or the semantic values themselves.
The remainder of this section assumes that no disambiguation strategies are in place.

As described next, without disambiguation strategies, the semantic phase has worst-case exponential running times as grammar exists with exponentially many derivations of sentences.
In fact, a grammar with a nonterminal that is both left- and right-recursive can yield infinitely many derivations.
However, to prevent non-termination, the semantic phase does not consider all the derivations of such \emph{cyclic nonterminals}.
A similar semantic phase was first described by Ridge~\cite{ridge2014}.

\begin{hscode}\SaveRestoreHook
\column{B}{@{}>{\hspre}l<{\hspost}@{}}%
\column{7}{@{}>{\hspre}l<{\hspost}@{}}%
\column{9}{@{}>{\hspre}l<{\hspost}@{}}%
\column{22}{@{}>{\hspre}c<{\hspost}@{}}%
\column{22E}{@{}l@{}}%
\column{E}{@{}>{\hspre}l<{\hspost}@{}}%
\>[B]{}\mathbf{type}\;{}\<[7]%
\>[7]{}\Conid{Evaluator}\;\Varid{t}\;\Varid{a}{}\<[22]%
\>[22]{}\mathrel{=}{}\<[22E]%
\\
\>[7]{}\hsindent{2}{}\<[9]%
\>[9]{}\Conid{Array}\;\Conid{Int}\;\Varid{t}\to \Conid{BSRs}\to \Conid{Int}\to \Conid{Int}\to \Conid{Set}\;\Conid{Symbol}\to [\mskip1.5mu \Varid{a}\mskip1.5mu]{}\<[E]%
\ColumnHook
\end{hscode}\resethooks
An evaluator function (third component of a \ensuremath{\Conid{Symbol}}) receives the input sentence \ensuremath{\Varid{inp}}, a set of BSR elements \ensuremath{\Varid{bsrs}}, a left extent \ensuremath{\Varid{l}}, a right extent \ensuremath{\Varid{r}} and a set of (nonterminal) symbols \ensuremath{\Varid{nts}}.
It returns a semantic value for every derivation of the subsentence of \ensuremath{\Varid{inp}} ranging from $l$ to $r-1$ encoded in \ensuremath{\Varid{bsrs}}, unless it is an evaluator for a nonterminal that is in \ensuremath{\Varid{nts}}.
The evaluator function for the nonterminal \ensuremath{\Varid{nt}} is defined in terms of the evaluator functions of the symbols occuring in the definition of \ensuremath{\Varid{nt}} (see the usage of \ensuremath{\Varid{eval}_\mathit{sm}} in Figure~\ref{fig_csv_code}).
The evaluator for \ensuremath{\Varid{nt}} might call itself recursively, as is the case for \ensuremath{\Conid{CSV}}.
It is possible to detect recursive calls, ensuring termination of the semantic phase, by maintaining a set of encountered nonterminals \ensuremath{\Varid{nts}} and seeing whether the evaluator for \ensuremath{\Varid{nt}} is called with \ensuremath{\Varid{nt}\in\Varid{nts}}.
In order to detect only that the evaluator is simultaneously left-recursive and right-recursive, the algorithm empties the set \ensuremath{\Varid{nts}} whenever \ensuremath{\Varid{l}} or \ensuremath{\Varid{r}} changes between two evaluator calls (see the definition of \ensuremath{\Varid{nts'}} in the example code).

The result of the evaluator for \ensuremath{\Varid{nt}} (\ensuremath{\Varid{res}} in Figure~\ref{fig_csv_code}) is the concatenation of the results for the alternates of \ensuremath{\Varid{nt}}.
If an alternate has $k >= 1$ symbols, then the evaluator finds all the $k$-length splits of the subsentence of \ensuremath{\Varid{inp}} ranging from $l$ to $r-1$ such that every $i$-th element of the split is matched by the $i$-th symbol of the alternate.
In essense, finding one or more such splits is what parsing algorithms do.
However, in this case, the information necessary to compute the splits is already encoded as the pivots in \ensuremath{\Varid{bsrs}}.
The function \ensuremath{\Varid{pivots}} (see Figure~\ref{fig_fungll_datas}) is given a triple \ensuremath{(\Varid{s},\Varid{l},\Varid{r})} and returns all the pivots $k$ such that \ensuremath{(\Varid{s},\Varid{l},\Varid{k},\Varid{r})\in\Varid{bsrs}}.
The slots $s$ are of the form $(x,\alpha y,\beta)$.
By producing pivot $k$, the BSR set informs the evaluator that \ensuremath{\alpha} matches with left extent $l$ and right extent $k$ and \ensuremath{\Varid{y}} matches with left extent $k$ and right $r$.
The evaluator finds all splits by continuing with the triple $((x,\alpha,y\beta),l,k)$ and so on for all $k$ and until all slots have been seen (except those of the form $(x,\epsilon,\beta)$).
For every split, the evaluator of the $i$-th symbol of the alternate is called with the left and right extents of the $i$-th element of the split.
The semantic values produced by these calls form the inputs of the semantic action that is attached to the alternate.

If $k = 0$, then a singleton list with the result of the semantic action is returned, or the empty list, if there is no BSR element $((\mathit{nt},\epsilon,\epsilon),l,l,r) \in \mathit{bsrs}$ (which can only be true if $l = r$).

The evaluator for a token (not shown) checks whether the given left and right extent are one apart, i.e. whether $l + 1 = r$, and if so returns the singleton list with the matched token in it (and the empty list otherwise).
\section{Implementation}
\label{sec_implementation}
The supplementary material of this paper contains a version of Happy that implements the GLL back-end.
Aspects of this implementation are discussed in this section.
The completeness with respect to the LALR back-end is discussed, as well as several avenues for improvements and extensions.

\paragraph{Datastructures and operations}
The previous section describes the essential data structures of the \fungll{} algorithm as sets, omitting their actual implementation and focusing on the method of generating parsers.
However, the efficiency and worst-case complexity of \fungll{} are strongly influenced by the datastructures and their operations.
A direct implementation as sets (e.g. from \ensuremath{\Conid{\Conid{Data}.Set}}) is inefficient. 
The type \ensuremath{\Conid{USet}} is therefore defined as nested integer tries~\cite{intmaps}, with a nesting-level for the two integers of descriptors:
\begin{hscode}\SaveRestoreHook
\column{B}{@{}>{\hspre}l<{\hspost}@{}}%
\column{E}{@{}>{\hspre}l<{\hspost}@{}}%
\>[B]{}\mathbf{type}\;\Conid{USet}\mathrel{=}\Conid{IntMap}\;(\Conid{IntMap}\;(\Conid{Set}\;\Conid{Slot})){}\<[E]%
\ColumnHook
\end{hscode}\resethooks
This approach is not sufficient however: the evaluation section shows that the GLL back-end is significantly slower than in the LALR and GLR back-ends when applied to LALR grammars.
The implementation work supporting this paper has been used to demonstrate the capabilities of the GLL back-end and to build up a collection of examples that can be used for testing faster implementations.
Alternative implementations may be explored in future work.
Faster parser generation strategies are also possible if modularity and reuse are not required.

\paragraph{Lookahead}
The GLL back-end does not currently implement a form of lookahead.
However, lookahead sets can be computed dynamically for every slot in a way similar to how the slot itself is computed.
A possible approach is to extend the type \ensuremath{\Conid{Symbol}} with lookahead sets and to generate expressions that computes these sets.
%
%
Lookahead sets can be computed statically if modularity and reuse are not required.

\paragraph{Monadic actions}
\begin{figure}
\begin{hscode}\SaveRestoreHook
\column{B}{@{}>{\hspre}l<{\hspost}@{}}%
\column{3}{@{}>{\hspre}l<{\hspost}@{}}%
\column{5}{@{}>{\hspre}l<{\hspost}@{}}%
\column{7}{@{}>{\hspre}l<{\hspost}@{}}%
\column{16}{@{}>{\hspre}l<{\hspost}@{}}%
\column{17}{@{}>{\hspre}l<{\hspost}@{}}%
\column{25}{@{}>{\hspre}l<{\hspost}@{}}%
\column{31}{@{}>{\hspre}l<{\hspost}@{}}%
\column{36}{@{}>{\hspre}c<{\hspost}@{}}%
\column{36E}{@{}l@{}}%
\column{39}{@{}>{\hspre}c<{\hspost}@{}}%
\column{39E}{@{}l@{}}%
\column{51}{@{}>{\hspre}c<{\hspost}@{}}%
\column{51E}{@{}l@{}}%
\column{52}{@{}>{\hspre}c<{\hspost}@{}}%
\column{52E}{@{}l@{}}%
\column{55}{@{}>{\hspre}c<{\hspost}@{}}%
\column{55E}{@{}l@{}}%
\column{E}{@{}>{\hspre}l<{\hspost}@{}}%
\>[B]{}\%\textbf{error}\;{}\<[16]%
\>[16]{}\{\mskip1.5mu \Varid{error}\mathbin{\circ}\Varid{show}\mskip1.5mu\}{}\<[E]%
\\
\>[B]{}\%\textbf{monad}\;{}\<[16]%
\>[16]{}\{\mskip1.5mu \Conid{SeedM}\;\Conid{Int}\mskip1.5mu\}{}\<[E]%
\\
\>[B]{}\%\textbf{token}\mathbin{...}{}\<[17]%
\>[17]{}\mbox{\onelinecomment  token directives have been elided }{}\<[E]%
\\
\>[B]{}\%\textbf{left}\;\text{\tt '-'}\;\text{\tt '+'}{}\<[17]%
\>[17]{}\mbox{\onelinecomment  order of \ensuremath{\%\textbf{left}} directives}{}\<[E]%
\\
\>[B]{}\%\textbf{left}\;\text{\tt '*'}\;\text{\tt '/'}{}\<[17]%
\>[17]{}\mbox{\onelinecomment  determines precedence}{}\<[E]%
\\
\>[B]{}\%\%{}\<[E]%
\\
\>[B]{}\Conid{Expr}{}\<[E]%
\\
\>[B]{}\hsindent{3}{}\<[3]%
\>[3]{}\mathbin{:}{}\<[7]%
\>[7]{}\Conid{Expr}\;\text{\tt '+'}\;\Conid{Expr}\;{}\<[25]%
\>[25]{}\{\mskip1.5mu \%{}\<[31]%
\>[31]{}\Varid{return}\;(\;\$1\mathbin{+}\;\$3){}\<[51]%
\>[51]{}\mskip1.5mu\}{}\<[51E]%
\\
\>[B]{}\hsindent{3}{}\<[3]%
\>[3]{}\mid {}\<[7]%
\>[7]{}\Conid{Expr}\;\text{\tt '-'}\;\Conid{Expr}\;{}\<[25]%
\>[25]{}\{\mskip1.5mu \%\%\;{}\<[31]%
\>[31]{}(\mathbin{-})\;\langle\$\rangle\;\;\$1\;\langle*\rangle\;\;\$3{}\<[52]%
\>[52]{}\mskip1.5mu\}{}\<[52E]%
\\
\>[B]{}\hsindent{3}{}\<[3]%
\>[3]{}\mid {}\<[7]%
\>[7]{}\Conid{Expr}\;\text{\tt '*'}\;\Conid{Expr}\;{}\<[25]%
\>[25]{}\{\mskip1.5mu {}\<[31]%
\>[31]{}\;\$1\mathbin{*}\;\$3\mskip1.5mu\}{}\<[E]%
\\
\>[B]{}\hsindent{3}{}\<[3]%
\>[3]{}\mid {}\<[7]%
\>[7]{}\Conid{Expr}\;\text{\tt '/'}\;\Conid{Expr}\;{}\<[25]%
\>[25]{}\{\mskip1.5mu \%\%\%\;{}\<[31]%
\>[31]{}[\mskip1.5mu \Varid{div}\;\langle\$\rangle\;\;\$1\;\langle*\rangle\;\;\$3{}\<[E]%
\\
\>[31]{}\mid \mathrm{0}\not\equiv \Varid{giveValue}\;\;\$3\mskip1.5mu]{}\<[55]%
\>[55]{}\mskip1.5mu\}{}\<[55E]%
\\
\>[B]{}\hsindent{3}{}\<[3]%
\>[3]{}\mid {}\<[7]%
\>[7]{}\text{\tt '('}\;\Conid{Expr}\;\text{\tt ')'}\;{}\<[25]%
\>[25]{}\{\mskip1.5mu {}\<[31]%
\>[31]{}\;\$2{}\<[36]%
\>[36]{}\mskip1.5mu\}{}\<[36E]%
\\
\>[B]{}\hsindent{3}{}\<[3]%
\>[3]{}\mid {}\<[7]%
\>[7]{}\Varid{digit}\;{}\<[25]%
\>[25]{}\{\mskip1.5mu {}\<[31]%
\>[31]{}\;\$1{}\<[36]%
\>[36]{}\mskip1.5mu\}{}\<[36E]%
\\
\>[B]{}\hsindent{3}{}\<[3]%
\>[3]{}\mid {}\<[7]%
\>[7]{}\text{\tt '\#'}\;{}\<[25]%
\>[25]{}\{\mskip1.5mu \%{}\<[31]%
\>[31]{}\Varid{enroll}{}\<[39]%
\>[39]{}\mskip1.5mu\}{}\<[39E]%
\\
\>[B]{}\{\mskip1.5mu {}\<[E]%
\\
\>[B]{}\mathbf{data}\;\Conid{SeedM}\;\Varid{s}\;\Varid{a}\mathrel{=}\Conid{SeedM}\;\{\mskip1.5mu \Varid{runSeedM}\mathbin{::}(\Varid{s}\to (\Varid{s},\Varid{a}))\mskip1.5mu\}{}\<[E]%
\\
\>[B]{}\mathbf{instance}\;\Conid{Monad}\;(\Conid{SeedM}\;\Varid{s})\;\mathbf{where}{}\<[E]%
\\
\>[B]{}\hsindent{3}{}\<[3]%
\>[3]{}\Varid{return}\;\Varid{a}\mathrel{=}\Conid{SeedM}\;(\lambda \Varid{i}\to (\Varid{i},\Varid{a})){}\<[E]%
\\
\>[B]{}\hsindent{3}{}\<[3]%
\>[3]{}(\Conid{SeedM}\;\Varid{p})\bind \Varid{mq}\mathrel{=}\Conid{SeedM}\mathbin{\$}\lambda \Varid{seed}\to {}\<[E]%
\\
\>[3]{}\hsindent{2}{}\<[5]%
\>[5]{}\mathbf{let}\;(\Varid{seed'},\Varid{pv})\mathrel{=}\Varid{p}\;\Varid{seed}{}\<[E]%
\\
\>[3]{}\hsindent{2}{}\<[5]%
\>[5]{}\mathbf{in}\;\Varid{runSeedM}\;(\Varid{mq}\;\Varid{pv})\;\Varid{seed'}{}\<[E]%
\\
\>[B]{}\Varid{giveValue}\mathbin{::}\Conid{SeedM}\;\Conid{Int}\;\Varid{a}\to \Varid{a}{}\<[E]%
\\
\>[B]{}\Varid{giveValue}\;\Varid{sm}\mathrel{=}\Varid{snd}\;(\Varid{runSeedM}\;\Varid{sm}\;\mathrm{1}){}\<[E]%
\\
\>[B]{}\Varid{enroll}\mathbin{::}\Conid{Enum}\;\Varid{s}\Rightarrow \Conid{SeedM}\;\Varid{s}\;\Varid{s}{}\<[E]%
\\
\>[B]{}\Varid{enroll}\mathrel{=}\Conid{SeedM}\;(\lambda \Varid{i}\to (\Varid{succ}\;\Varid{i},\Varid{i})){}\<[E]%
\\
\>[B]{}\mskip1.5mu\}{}\<[E]%
\ColumnHook
\end{hscode}\resethooks

\caption{Example grammar demonstrating additional Happy features implemented by the GLL back-end.}
\label{fig_expr_example}
\end{figure}

Figure~\ref{fig_expr_example} shows a Happy grammar defining a small expression language for integer arithmetic by using several additional features of Happy implemented by the GLL back-end.
The semantic actions define an interpreter for this language.
The \ensuremath{\%\textbf{monad}} directive influences the type signatures of the generated functions as well as the treatment of semantic actions.
In the example, the semantic values produced by the semantic actions are computations in the \ensuremath{\Conid{SeedM}} monad, as specified by the \ensuremath{\%\textbf{monad}} directive.
The language is defined such that whenever a `\#' is encountered, a fresh integer is generated by applying \ensuremath{\Varid{enroll}}.

The example shows four kinds of semantic actions, distinguished by the zero, one, two or three percentage symbols that precede the action.
The first kind \ensuremath{\{\mskip1.5mu \mathbin{...}\mskip1.5mu\}} is used to write semantic actions in the same way with or without \ensuremath{\%\textbf{monad}} directive (e.g. the alternate for multiplication).
Since there is a \ensuremath{\%\textbf{monad}} directive in the example, the action is transformed before it is inserted in the generated parser, to form an equivalent of:
\begin{hscode}\SaveRestoreHook
\column{B}{@{}>{\hspre}l<{\hspost}@{}}%
\column{6}{@{}>{\hspre}l<{\hspost}@{}}%
\column{E}{@{}>{\hspre}l<{\hspost}@{}}%
\>[B]{}\mathbf{do}\;{}\<[6]%
\>[6]{}\Varid{v\char95 1}\leftarrow \Varid{x\char95 1}{}\<[E]%
\\
\>[6]{}\Varid{v\char95 2}\leftarrow \Varid{x\char95 2}{}\<[E]%
\\
\>[6]{}\Varid{v\char95 3}\leftarrow \Varid{x\char95 3}{}\<[E]%
\\
\>[6]{}\Varid{return}\;(\Varid{v\char95 1}\mathbin{*}\Varid{v\char95 3})\mbox{\onelinecomment  for the action \ensuremath{\{\mskip1.5mu \;\$1\mathbin{*}\;\$3\mskip1.5mu\}}}{}\<[E]%
\ColumnHook
\end{hscode}\resethooks
Informally, the three monadic parse results for the three symbols of the alternate are `run' to yield their values.
The semantic action written by the programmer is the returned expression, after replacing $\$1$ and $\$3$ with the identifiers binding the semantic values of the first and third symbol.
The second kind \ensuremath{\{\mskip1.5mu \%\mathbin{...}\mskip1.5mu\}} of semantic action differs in that there is no implicit application of \ensuremath{\Varid{return}}; the user expression must be of the right monadic type.
This means that the programmer has access to the monad in the semantic action (e.g. see \ensuremath{\Varid{enroll}} in the alternate for \ensuremath{\text{\tt '\#'}}).
In the third kind \ensuremath{\{\mskip1.5mu \%\%\mathbin{...}\mskip1.5mu\}}, the action parameters $\$1$, $\$2$, $\$3$, etc. are replaced by the identifiers binding the monadic parse results of the alternate's symbols without `running' them.
The programmer thus has complete control over the monadic computation constructed by the semantic action (e.g. see the alternate for subtraction).
On top of this, the fourth\footnote{If there is no \ensuremath{\%\textbf{monad}} directive, the behavior of the fourth kind is given to the second kind of semantic action.} kind \ensuremath{\{\mskip1.5mu \%\%\%\mathbin{...}\mskip1.5mu\}} gives the programmer access to the list-monad of the evaluation phase.
This powerful feature can be used to implement arbitrary disambiguation strategies based on the semantic values of sub-expressions.
In the example, this feature is used to rule out expressions in which a division by zero error would occur.

\paragraph{Error handlers}
The example also shows that the \ensuremath{\%\textbf{error}} directive can be used to specify a handler for parse errors.
This is not trivial, because the GLL algorithm explores all possible interpretations of an input sentence.
In the case of a parse failure, there are likely several points of failure and it is not clear which should be reported as errors.
A wrapper function is generated for each production of a grammar that applies the parser for the production and yields errors as the left component of an \ensuremath{\Conid{Either}}.
There are $n$ error values\footnote{Their type is inferred from the error handler.}, one for each of the $n$ parsing attempts reaching the furthest into the input sentence.
The number $n$ is determined by a configuration option during parser generation or at runtime.
%
%

\paragraph{Disambiguation}
The disambiguation directives \ensuremath{\%\textbf{left}}, \ensuremath{\%\textbf{right}} and \ensuremath{\%\textbf{nonassoc}} behave slightly different in the GLL back-end than the LALR and GLR back-ends.
This is because the GLL back-end implements the directives by filtering BSR sets rather than resolving conflicts in a parse table.
The precise difference in behavior is to be investigated further.

\paragraph{Threaded lexers}
The \ensuremath{\%\textbf{lexer}} directive of Happy makes it possible to parse based on a `threaded lexer monad' that propagates the input sentence in its state.
This enables error handlers and semantic actions that interact with the lexer state, for example to produce messages based on the line and column number of the next character in the input.
This feature has not been implemented.
An implementation would require the parsing algorithm to be able to `reset' lexer state in order to explore the possibly many ways a sentence can be parsed.

\paragraph{Attribute grammars}

The semantic actions of Happy allow the programmer to write arbitrary syntax-directed translations in the style of `The Dragon Book'~\cite{dragon_book}. 
The Attribute Grammar formalism~\cite{knuth68} can be explained as a particular way of writing syntax-directed translations in which semantic \emph{equations} define attributes in terms of each other.
The equations are solved by an evaluator that traverses a program by `visiting' its components possibly many times until all attributes are assigned a value.
Haskell's lazy-evaluation makes it possible to generate evaluation functions for each production of the attribute grammar in a modular fashion~\cite{saraiva99,swierstra1999designing}.
The evaluator fails to terminate, however, if the dependencies between attributes form a cycle.
Happy employs this strategy and implements attribute grammars as an alternative to semantics actions.
There should be no problem making attribute grammar evaluation available to the GLL back-end as Happy's evaluation strategy for attribute grammars is a natural fit with the semantic phase of the generated GLL parsers.
However, the current implementation does not yet demonstrate this.
%
%

\section{Evaluation}
\label{sec_evaluation}
%
%

%
This section evaluates characteristics of the GLL back-end implementation in comparison to the LALR and GLR back-ends.
%
%
%
%
%
The modularity and compositionality aspects that are unique to the GLL back-end have been discussed in Section~\ref{sec_motivation}.
All experiments have been executed under Linux mint 20.3 on a laptop with an Intel i7-8565U (8) @ 4.600GHz CPU and 16GiB of RAM using version 8.6.5 of the Glasgow Haskell Compiler (GHC).
The tools and files necessary to reproduce the results of the experiments are provided as supplementary material.

%

\paragraph{Beyond context-free grammars}
Happy's parameterized nonterminals can be used to write grammars for languages that are not context-free.
For example, the following code fragment shows a grammar for the language $a^nb^nc^n$ with $n \geq 1$ (this example is taken from~\cite{fischer68}).
\begin{hscode}\SaveRestoreHook
\column{B}{@{}>{\hspre}l<{\hspost}@{}}%
\column{11}{@{}>{\hspre}c<{\hspost}@{}}%
\column{11E}{@{}l@{}}%
\column{14}{@{}>{\hspre}l<{\hspost}@{}}%
\column{E}{@{}>{\hspre}l<{\hspost}@{}}%
\>[B]{}\Conid{Start}{}\<[11]%
\>[11]{}\mathbin{:}{}\<[11E]%
\>[14]{}\Conid{F}\;(\text{\tt 'a'},\text{\tt 'b'},\text{\tt 'c'}){}\<[E]%
\\
\>[B]{}\Conid{F}\;(\Varid{x},\Varid{y},\Varid{z}){}\<[11]%
\>[11]{}\mathbin{:}{}\<[11E]%
\>[14]{}\Conid{F}\;(\Conid{Seq}\;(\Varid{x},\text{\tt 'a'}),\Conid{Seq}\;(\Varid{y},\text{\tt 'b'}),\Conid{Seq}\;(\Varid{z},\text{\tt 'c'})){}\<[E]%
\\
\>[11]{}\mid {}\<[11E]%
\>[14]{}\Varid{x}\;\Varid{y}\;\Varid{z}{}\<[E]%
\\
\>[B]{}\Conid{Seq}\;(\Varid{p},\Varid{q}){}\<[11]%
\>[11]{}\mathbin{:}{}\<[11E]%
\>[14]{}\Varid{p}\;\Varid{q}{}\<[E]%
\ColumnHook
\end{hscode}\resethooks
%
%
%
%
The LALR and GLR back-ends depend on a fix-point algorithm that replaces parameterized nonterminals with specialized (non-parameterized) variants in a process similar to macro-expansion.
However, this algorithm cannot be used to remove parameterized nonterminals that apply themselves recursively with arguments that change with every recursive call.
In this example, the algorithm tries to compute the grammar that has infinitely many specializations of \ensuremath{\Conid{F}}, one for every choice of $n$.
Happy does not detect this and fails to terminate on this example without warning.
%
%

As discussed in Sections~\ref{sec_motivation} and~\ref{sec_fungll}, the GLL back-end generates parsers that take advantage of Haskell's abstraction mechanism to implement parameterized nonterminals so that applications are executed dynamically.
In the case of this example, a GLL parser is generated, but it fails to terminate on any input sentence.
This is because each recursive call to the parser implementing \ensuremath{\Conid{F}} produces a fresh nonterminal name and thus a descriptor that is unique to the call and therefore not already in the descriptor set.
This can be seen as a higher-order variant of the problem of left-recursion.

The grammar in the following fragment generates the context-sensitive language $\sum_{i=0}^{\infty}(b^iac^i)$. That is, the language \ensuremath{\{\mskip1.5mu \text{\tt \char34 a\char34},\text{\tt \char34 a(a)\char34},\text{\tt \char34 a(a)((a))\char34},\mathbin{...}\mskip1.5mu\}} with \ensuremath{\text{\tt '('}} instead of \ensuremath{\text{\tt 'b'}} and \ensuremath{\text{\tt ')'}} instead of \ensuremath{\text{\tt 'c'}} for clarity.
\begin{hscode}\SaveRestoreHook
\column{B}{@{}>{\hspre}l<{\hspost}@{}}%
\column{15}{@{}>{\hspre}c<{\hspost}@{}}%
\column{15E}{@{}l@{}}%
\column{19}{@{}>{\hspre}l<{\hspost}@{}}%
\column{E}{@{}>{\hspre}l<{\hspost}@{}}%
\>[B]{}\Conid{Start}{}\<[15]%
\>[15]{}\mathbin{:}{}\<[15E]%
\>[19]{}\Conid{List}\;(\text{\tt 'a'}){}\<[E]%
\\
\>[B]{}\Conid{List}\;(\Varid{e}){}\<[15]%
\>[15]{}\mathbin{:}{}\<[15E]%
\>[19]{}\Varid{e}{}\<[E]%
\\
\>[15]{}\mid {}\<[15E]%
\>[19]{}\Varid{e}\;\Conid{List}\;(\Conid{Parens}\;(\Varid{e})){}\<[E]%
\\
\>[B]{}\Conid{Parens}\;(\Varid{e}){}\<[15]%
\>[15]{}\mathbin{:}{}\<[15E]%
\>[19]{}\text{\tt '('}\;\Varid{e}\;\text{\tt ')'}{}\<[E]%
\ColumnHook
\end{hscode}\resethooks
The recursion of \ensuremath{\Conid{List}} is on the right and the GLL parser generated for this grammar indeed recognizes the language.
This is because every recursive call to the parser for \ensuremath{\Conid{List}} receives an index into the input sentence that is closer towards the end of the sentence.
No further recursive calls are made once the whole sentence is `consumed'.
This observation suggests that a variant of the curtailment procedure of~\cite{frost2008} can overcome the problem of higher-order left-recursion by making at most as many recursive calls as there are tokens left in the input sentence.
The other back-ends cannot handle this example for the reason mentioned before.

\paragraph{Permutation phrases}
The next experiment is about parsing `permutation phrases'~\cite{baars2004b}.
A similar experiment has been performed in~\cite{binsbergen2020}.
A permutation phrase is a sequence of permutable elements in which each element occurs exactly once and in any order~\cite{cameron93}.
In the permutation phrases considered here, elements occur \emph{at most} once.
Real-world examples of such permutation phrases are the modifiers associated with fields and methods in Java~\cite{thiemann2008} and the declaration specifiers of C~\cite{cameron93}.
The syntax of permutation phrases can be captured by a nonterminal with an alternate for each of the possible permutations.
Such a grammar is not practical as the number of alternates grows exponentially with the number of permutable elements.
The following fragment shows how the syntax of permutation phrases is captured conveniently with parameterized nonterminals.
This formulation is based on the \texttt{PermP3} example of~\cite{thiemann2008}.
The permutable elements are the digits 1 to 4.
The character \ensuremath{\text{\tt '\$'}} is assumed not to occur in any input sentence, thus ensuring that the parser for \ensuremath{\Conid{Nul}} always fails.
Each alternate of \ensuremath{\Conid{Choose}} chooses one of the elements and makes a recursive call to continue choosing.
In the recursive call, the chosen element is replaced with \ensuremath{\Conid{Nul}} so that it can no longer be chosen.
\begin{hscode}\SaveRestoreHook
\column{B}{@{}>{\hspre}l<{\hspost}@{}}%
\column{17}{@{}>{\hspre}l<{\hspost}@{}}%
\column{18}{@{}>{\hspre}l<{\hspost}@{}}%
\column{21}{@{}>{\hspre}l<{\hspost}@{}}%
\column{43}{@{}>{\hspre}l<{\hspost}@{}}%
\column{E}{@{}>{\hspre}l<{\hspost}@{}}%
\>[B]{}\Conid{Permutations}\mathbin{:}{}\<[17]%
\>[17]{}\Conid{Choose}\;(\text{\tt '1'},\text{\tt '2'},\text{\tt '3'},\text{\tt '4'})\;{}\<[43]%
\>[43]{}\{\mskip1.5mu \;\$1\mskip1.5mu\}{}\<[E]%
\\
\>[B]{}\Conid{Choose}\;(\Varid{a},\Varid{b},\Varid{c},\Varid{d}){}\<[18]%
\>[18]{}\mathbin{:}{}\<[43]%
\>[43]{}\{\mskip1.5mu [\mskip1.5mu \mskip1.5mu]\mskip1.5mu\}{}\<[E]%
\\
\>[18]{}\mid {}\<[21]%
\>[21]{}\Varid{a}\;\Conid{Choose}\;(\Conid{Nul},\Varid{b},\Varid{c},\Varid{d})\;{}\<[43]%
\>[43]{}\{\mskip1.5mu (\;\$1\mathbin{:}\;\$2)\mskip1.5mu\}{}\<[E]%
\\
\>[18]{}\mid {}\<[21]%
\>[21]{}\Varid{b}\;\Conid{Choose}\;(\Varid{a},\Conid{Nul},\Varid{c},\Varid{d})\;{}\<[43]%
\>[43]{}\{\mskip1.5mu (\;\$1\mathbin{:}\;\$2)\mskip1.5mu\}{}\<[E]%
\\
\>[18]{}\mid {}\<[21]%
\>[21]{}\Varid{c}\;\Conid{Choose}\;(\Varid{a},\Varid{b},\Conid{Nul},\Varid{d})\;{}\<[43]%
\>[43]{}\{\mskip1.5mu (\;\$1\mathbin{:}\;\$2)\mskip1.5mu\}{}\<[E]%
\\
\>[18]{}\mid {}\<[21]%
\>[21]{}\Varid{d}\;\Conid{Choose}\;(\Varid{a},\Varid{b},\Varid{c},\Conid{Nul})\;{}\<[43]%
\>[43]{}\{\mskip1.5mu (\;\$1\mathbin{:}\;\$2)\mskip1.5mu\}{}\<[E]%
\\
\>[B]{}\Conid{Nul}{}\<[18]%
\>[18]{}\mathbin{:}\text{\tt '\$'}\;{}\<[43]%
\>[43]{}\{\mskip1.5mu \text{\tt '\$'}\mskip1.5mu\}{}\<[E]%
\ColumnHook
\end{hscode}\resethooks

\begin{table}[t]%
\begin{tabular}{|l|r|r|r|r|r|}%
\hline
	\textbf{Alg.} & \# & \textbf{Generate} & \textbf{Size} & \textbf{Compile} & \textbf{Parse} \\
\hline
LALR & 4 & 0.009s& 45.0KiB& 1.226s& 0.001s \\
& 5 & 0.054s& 88.0KiB& 3.003s& 0.001s \\
& 6 & 0.679s& 191.0KiB& 11.751s& 0.001s \\
\hline
GLL & 4 & 0.004s& 34.0KiB& 0.964s& 0.001s \\
& 5 &  0.004s& 35.0KiB& 0.983s& 0.002s \\ 
& 6 & 0.004s& 37.0KiB& 1.026s& 0.002s \\
\hline
\end{tabular}%
\caption{Parsing permutation phrases.}%
\label{tab_permutations}%
\end{table}
Happy's algorithm for removing parameterized nonterminals successfully generates an equivalent grammar for this example.
However, the resulting grammar has exponentially many alternates relative to the number of permutable elements.
%
%
Table~\ref{tab_permutations} demonstrates the exponential growth by showing data about the LALR parsers generated for the syntax of permutation phrases with four, five and six elements.
The table shows the time it took to generate the parser, the size of the generated parser in kilobytes, the time it took GHC to compile this parser and the time it took the parser to parse a permutation.
The data for the GLL parsers generated from the same grammars, without removing parameterized nonterminals, show a small linear growth instead. 

%
\paragraph{LALR grammars}
In the next experiment, the running times of the parsers generated by the three different back-ends are compared with LALR grammars.
The BNF Convertor (BNFC) generates lexers, parsers and abstract syntax from a single grammar description written in the Labelled BNF (LBNF) formalism~\cite{bnfc_online}.
In fact, the convertor generates a small pipeline that runs the lexer and parser, creates an abstract syntax tree and pretty-prints it.
The BNFC tool is capable of generating two types of Happy grammars from an LBNF grammar, one for the LALR back-end -- to which the GLL back-end can also be applied -- and another specialized for the GLR back-end, making it the perfect tool for generating the inputs of this experiment.
The BNFC project includes a substantial amount of LBNF grammars for real-world languages such as Java, ANSI-C and Prolog.
However, the examples do not come with many tests and it is not always clear for which version of the language the syntax has been described.
Experiments have been performed with pipelines for ANSI-C and for LBNF itself.

\begin{table}
\begin{tabular}{|l|r|r|r|r|r|r|}%
\hline
\textbf{Alg.} &194&424&934&1541&2233\\
\hline
LALR& 0.0s& 0.0s& 0.01s& 0.01s& 0.01s\\
GLR& 0.01s& 0.02s& 0.04s& 0.06s& 0.08s\\
GLL& 0.02s& 0.04s& 0.11s& 0.17s& 0.29s\\
\hline
\end{tabular}%
\caption{Running times for LBNF pipelines.}%
\label{tab_bnfc_experiments_bnfc}%
\end{table}
The experiment with the LBNF pipeline has been conducted with the LBNF grammars for Prolog, LBNF itself, OCL (Object Constraint Language), GF (Grammatical Framework) and ANSI-C as test inputs.
Table~\ref{tab_bnfc_experiments_bnfc} shows the running times of executing the pipeline on these inputs.
The column headers contain the number of tokens produced by the lexer of the pipeline.
The only difference between the LALR, GLR and GLL rows of the table are the back-end used to generate the parser of the pipeline.

\begin{table}[t]
\begin{tabular}{|l|r|r|r|r|r|r|}%
\hline
\textbf{Alg.}&46&271&7261&15425(x2.1)&26106(x1.7)\\
\hline
LALR& 0.0s& 0.0s& 0.05s& 0.1s(x2.0)& 0.18s(x1.8)\\
GLR& 0.01s& 0.02s& 0.57s& 1.23s(x2.2)& 2.0s(x1.6)\\
GLL& 0.04s& 0.23s& 7.13s& 15.71s(x2.2)& 28.41s(x1.8)\\
\hline
\end{tabular}%
\caption{Running times for ANSI-C pipelines.}%
\label{tab_bnfc_experiments_c}%
\end{table}
The experiment with the ANSI-C pipeline has been conducted in a similar way, with two small programs and one large program taken from the BNFC repository.
The largest program is large indeed, with over 7500 lines of code and 187KB in size.
Two smaller, but still large, programs have been constructed by copying to separate files the first 2500 and 5000 lines of code of the largest program.
This has been done to demonstrate linear growth in running times.
Table~\ref{tab_bnfc_experiments_c} shows the running times of executing the pipeline for ANSI-C on the five input programs.
The last and second to last columns have multipliers indicating the growth of the value in the cell with respect to the previous column.

Tables~\ref{tab_bnfc_experiments_bnfc} and~\ref{tab_bnfc_experiments_c} show that the LALR parsers are significantly faster than the GLR parsers which in turn are significantly faster than the GLL parsers on LALR grammars.
The implementation of the GLL back-end has not yet been optimized for performance.
Various possible approaches to speeding up the GLL parsers have been discussed in the previous section.
For example, lookahead sets can be computed dynamically (to preserve modularity).
Alternatively, GLL parsers can be generated without a concern for modularity and parameterized nonterminals, thereby allowing more optimizations such as statically computed lookahead sets. 

\paragraph{Highly ambiguous grammars}
In the next experiment, the GLL and GLR back-ends are tested on highly ambiguous grammars.
The chosen grammars have been used by Ridge~\cite{ridge2014} in a comparison with the GLR back-end of Happy.
The grammars form a significant stress-test for complete parsers.
The LALR back-end is not used in the experiment because it does not produce complete parsers.
Each grammar consists of a single non-terminal and generates the language $a^n$ with $n\geq 0$.
The nonterminals are defined as follows:

\begin{minipage}{.32\columnwidth}
\begin{hscode}\SaveRestoreHook
\column{B}{@{}>{\hspre}l<{\hspost}@{}}%
\column{5}{@{}>{\hspre}c<{\hspost}@{}}%
\column{5E}{@{}l@{}}%
\column{8}{@{}>{\hspre}l<{\hspost}@{}}%
\column{E}{@{}>{\hspre}l<{\hspost}@{}}%
\>[B]{}S_1{}\<[5]%
\>[5]{}\mathbin{:}{}\<[5E]%
\>[8]{}\text{\tt 'a'}\;S_1\;S_1{}\<[E]%
\\
\>[5]{}\mid {}\<[5E]%
\ColumnHook
\end{hscode}\resethooks
\end{minipage}%
\begin{minipage}{.32\columnwidth}%
\begin{hscode}\SaveRestoreHook
\column{B}{@{}>{\hspre}l<{\hspost}@{}}%
\column{5}{@{}>{\hspre}c<{\hspost}@{}}%
\column{5E}{@{}l@{}}%
\column{8}{@{}>{\hspre}l<{\hspost}@{}}%
\column{E}{@{}>{\hspre}l<{\hspost}@{}}%
\>[B]{}S_2{}\<[5]%
\>[5]{}\mathbin{:}{}\<[5E]%
\>[8]{}S_2\;S_2\;\text{\tt 'a'}{}\<[E]%
\\
\>[5]{}\mid {}\<[5E]%
\ColumnHook
\end{hscode}\resethooks
\end{minipage}
\begin{minipage}{.32\columnwidth}
\begin{hscode}\SaveRestoreHook
\column{B}{@{}>{\hspre}l<{\hspost}@{}}%
\column{5}{@{}>{\hspre}c<{\hspost}@{}}%
\column{5E}{@{}l@{}}%
\column{8}{@{}>{\hspre}l<{\hspost}@{}}%
\column{E}{@{}>{\hspre}l<{\hspost}@{}}%
\>[B]{}\Conid{E}{}\<[5]%
\>[5]{}\mathbin{:}{}\<[5E]%
\>[8]{}\Conid{E}\;\Conid{E}\;\Conid{E}{}\<[E]%
\\
\>[5]{}\mid {}\<[5E]%
\>[8]{}\text{\tt 'a'}{}\<[E]%
\\
\>[5]{}\mid {}\<[5E]%
\ColumnHook
\end{hscode}\resethooks
\end{minipage} 

\noindent
The grammars are such that the number of derivations grows exponentially relative to the size of the input.
Efficiently \emph{enumerating} all derivations is therefore not possible and the GLR and GLL parsers are used for recognition only.
The GLL parsers still compute a BSR set that embeds all derivations. 

\begin{table}%
\begin{tabular}{|l|l|r|r|r|r|r|r|r|}%
\hline
\textbf{Alg.} & \textbf{Nt.} & 20 & 30 & 40 & 50 & 100 & 200 \\
\hline
GLR& $S_1$ & 0.11s& 0.81s& 4.18s& 13.96s& -& -\\
& $S_2$ & 0.004s& 0.01s& 0.02s& 0.03s& 0.32s& 3.46s\\
& E & 0.17s& 1.73s& 8.18s& 34.96s& -& -\\
\hline
GLL& $S_1$ & 0.006s& 0.02s& 0.05s& 0.08s& 0.82s& 7.43s\\
& $S_2$ & 0.007s& 0.01s& 0.03s& 0.06s& 0.45s& 3.1s\\
& E & 0.012s& 0.03s& 0.08s& 0.12s& 1.06s& 7.72s\\
\hline
\end{tabular}%
\caption{Recognition times for highly ambiguous grammars.}%
\label{tab_ridge_experiments}%
\end{table}
Table~\ref{tab_ridge_experiments} shows the running times of the GLR and GLL parsers given inputs that contain between 20 to 200 repetitions of the character \ensuremath{\text{\tt 'a'}}.
The GLL parsers for these grammars are significantly faster than the GLR parsers.
In fact, the GLR parsers for nonterminals $S_1$ and $E$ seem to suffer from exponential blow up (the GLR parsers for $S_1$ and $E$ ran unsuccessfully with a timeout of 500 seconds on 100 tokens) although the GLR algorithm theoretically does not~\cite{tomita1985,scott2007}.
\section{Related work}
\label{sec_related_work}
The GLL algorithm is a relatively new addition to the parsing landscape~\cite{scott2010,scott2013}, with several variations~\cite{afroozeh2015,scott2016,binsbergen2018a,binsbergen2020}, that has rekindled the interest in top-down parsing.
Implementations of the algorithm are not yet widespread, but can be found in parser generators~\cite{scott2016,basten2015} and combinator libraries~\cite{afroozeh2016,binsbergen2018a,binsbergen2020}.
To our knowledge, the GLL back-end for Happy is the first parser generator that combines GLL parsing with a facility for abstraction and reuse, benefitting directly from the top-down nature of the algorithm.
The \precc{} compiler generator combines LL($\infty$) parsing with a facility for abstraction and reuse, but does not generate complete parsers~\cite{precc}.
The OCaml parser generator Menhir generates LR(1) parsers based on grammars with parameterized rules and includes a warning for rules with unbounded growth that cause non-termination~\cite{menhir}.
Menhir offers a library of reusable rules for optionality, sequences and lists.
As embedded domain-specific languages, parser combinator libraries~\cite{swierstra2009} and grammar combinator libraries~\cite{devriese2011} support reuse naturally but vary wildly in the class of languages they accept and by the ease with which parsers are written.
For a lengthier discussion on this topic, the reader is referred to~\cite{binsbergen_thesis,binsbergen2020}.

Introduced by Fischer in 1968~\cite{fischer68}, macro-grammars extend context-free grammars by introducing `macro-like productions', similar to the parameterized nonterminal definitions of Happy.
An argument of a macro is a sequence of symbols, whereas an argument of a parameterized nonterminal is a single symbol.
So although macro-grammars are more expressive, it should be possible to write a Happy grammar for every macro-grammar in the way demonstrated by the nonterminal \ensuremath{\Conid{F}} in Section~\ref{sec_evaluation}.
Thiemann and Neubauer discuss generating LR parsers for restricted macro-grammars~\cite{thiemann2004} and describe an algorithm for checking whether a macro-grammar can be transformed into a context-free grammar~\cite{thiemann2008}.
A variant of this algorithm can be implemented in Happy to prevent non-termination of the algorithm that removes parameterized nonterminals for the LALR and GLR back-ends.
Perhaps this algorithm can be extended to detect what was called `higher-order left-recursion' in Section~\ref{sec_evaluation} in order to prevent the GLL back-end from generating parsers for parameterized nonterminals that fail to terminate.

\section{Conclusion}
\label{sec_conclusion}
This paper has presented a strategy for generating modular, reusable and complete top-down parsers from syntax descriptions with parameterized nonterminals, based on the \fungll{} variant of the GLL algorithm.
The strategy has been implemented in Haskell as a new back-end for Happy.
The ideas in this paper should be transferable to other grammar formalisms and host languages with higher-order functions.
The generated parsers are easy to test and debug because each grammar symbol is directly implemented as an executable parse function.
Moreover, the GLL back-end supports parameterization directly by generating higher-order parse functions reminiscent of parser combinators.
As a result, the back-end can generate practical parsers for a large class of grammars, including all context-free grammars and certain grammars that describe context-sensitive languages.
%
%

The new Happy back-end has been developed as a practical alternative to the LALR and GLR back-ends, whilst extending the functionality and usability of Happy by inheriting the positive aspects of recursive descent parsing, such as modularity, and realizing the full potential of `reuse through abstraction'.
%
%
The runtime efficiency of the generated GLL parsers can be improved, however.
The GLL parsers are significantly faster than the GLR parsers for highly ambiguous grammars but significantly slower than the LALR and GLR parsers for LALR grammars.
The running times of the generated GLL parsers can perhaps be improved by reimplementing the datastructures provided by the support library and by including lookahead tests.
However, some of the envisioned efficiency improvements require precomputing information based on the grammar as a whole, resulting in parsers that are not reusable across files and projects.
These possible runtime improvements are to be explored in future work.

%

\bibliographystyle{ACM-Reference-Format}
\bibliography{biblio}
\end{document}

%% file: abbrevs.tex
\newcommand{\precc}{\texttt{PRECC}}

%% file: abstract.tex
Parser generators and parser combinator libraries are the most popular tools for producing parsers.
Parser combinators use the host language to provide reusable components in the form of higher-order functions with parsers as parameters.
Very few parser generators support this kind of reuse through abstraction and even fewer generate parsers that are as modular and reusable as the parts of the grammar for which they are produced.
This paper presents a strategy for generating modular, reusable and complete top-down parsers from syntax descriptions with parameterized nonterminals, based on the FUN-GLL variant of the GLL algorithm.

The strategy is discussed and demonstrated as a novel back-end for the Happy parser generator.
Happy grammars can contain `parameterized nonterminals' in which parameters abstract over grammar symbols, granting an abstraction mechanism to define reusable grammar operators.
However, the existing Happy back-ends do not deliver on the full potential of parameterized nonterminals as parameterized nonterminals cannot be reused across grammars. 
Moreover, the parser generation process may fail to terminate or may result in exponentially large parsers generated in an exponential amount of time.

The GLL back-end presented in this paper implements parameterized nonterminals successfully by generating higher-order functions that resemble parser combinators, inheriting all the advantages of top-down parsing.
The back-end is capable of generating parsers for the full class of context-free grammars, generates parsers in linear time and generates parsers that find all derivations of the input string.
To our knowledge, the presented GLL back-end makes Happy the first parser generator that combines all these features.

This paper describes the translation procedure of the GLL
back-end and compares it to the LALR and GLR
back-ends of Happy in several experiments.

%% file: introduction.tex
\section{Introduction}
\label{sec_introduction}

\textbf{Recursive descent parsing} is a technique for (manually or mechanically) writing \emph{top-down} parsers based on the description of a context-free grammar. 
Recursive descent parsers have in common that every (nonterminal and terminal) symbol of the grammar is implemented by a piece of code, that these pieces of code can be placed in a sequence -- implementing an alternate of the grammar -- and that the choice between a nonterminal's alternates is implemented by branching control-flow.
Recursive descent parser generators implement a direct translation from grammar to parser.
Every symbol is translated independently (separate compilation) and the code for symbols can be tested independently (semantic modularity).
As a result, generated parsers are easy to maintain and debug; updated grammar specifications require minimal recompilation and unexpected behavior can be identified by isolating the parts of the parser and input that cause the unexpected behavior.

The functional characteristics of recursive descent parsers vary depending on the top-down parsing algorithm on which they are based, but recursive descent parsers have historically struggled with left-recursive nonterminals and non-factorized alternates.
Typical implementations only accept LL(k) grammars: the set of context-free grammars for which it holds that, with $k \geq 0$ terminal symbols lookahead, no two alternates are simultaneously applicable.
Employing a recursive descent parser generator often involves applying grammar transformations to remove left-recursion and/or to apply left-factorization to produce an LL(k) grammar.
However, applying grammar transformations may not always be desirable or even possible; there are context-free grammars for which there is no LL(k) equivalent.

\textbf{Bottom-up parsing} is more often possible without applying grammar transformations.
Bottom-up parsers are also very fast, benefiting from pre-computed information recorded in a table -- the so-called `parse table'.
These properties have made bottom-up parsing more popular than top-down parsing as the basis for parser generators, as evidenced by the large variety of algorithms such as LALR, SLR and GLR~\cite{tomita1985} and implementations such as by Yacc, Bison, Happy~\cite{happy}, Menhir~\cite{menhir}, Rascal~\cite{rascal2009}, Spoofax~\cite{kats2010} and SDF~\cite{vandenbrand2002}.
However, the users of bottom-up parser generators do not benefit from separate compilation and semantic modularity.
This paper discusses a new back-end for Happy which generates recursive descent parsers based on the generalized LL (GLL) algorithm.
The back-end has the aforementioned advantages of recursive descent parsing, but does not require any grammar transformations.
\textbf{Generalized parsing algorithms} are \textit{general} in the sense that they accept arbitrary context-free grammars and \textit{complete} in the sense that they produce all possible derivations of a given input string.
The generalized LL (GLL) algorithm~\cite{scott2013,scott2016} relies on intricate bookkeeping within potentially large datastructures to simultaneously ensure that parsers terminate and find all derivations. 
Despite this added complexity, GLL parsers are still easy to maintain, debug and support separate compilation and semantic modularity (like other recursive descent parsers).

\textbf{Generalized parser combinators} are recent technologies that combine the benefits of generalized parsing and parser combinators~\cite{ridge2014,afroozeh2016,binsbergen2018a,binsbergen2020}.
The mentioned approaches have in common that they involve explicit representations of grammar components such as symbols and productions rules -- as is required by Earley's generalized parsing algorithm~\cite{earley1970}, GLR~\cite{tomita1985} and GLL~\cite{scott2013} -- whereas conventional parser combinators have no explicit representation of grammar components.
The libraries presented by~\cite{ridge2014} and~\cite{binsbergen2018a} generate an actual grammar object before providing it as input to a standalone generalized parsing algorithm (voiding separate compilation). 
This idea of so-called grammar combinators has also been applied outside the context of generalized parsing by \cite{ljunglof2002}, \cite{baars2004}, and \cite{devriese2011}.

\textbf{Grammar combinators} blur the line between combinator libraries and parser generators, leaving only a few essential differences: grammar combinator libraries generate parsers at runtime rather than in a separate phase and grammar combinator libraries define \emph{embedded} domain-specific languages (EDSLs) whereas parser generators define external domain-specific languages (DSLs).
EDSLs are typically easy to extend, with the power of the host language available to define new operators. 
This holds true for parser combinators and, to a lesser degree, for grammar combinators~\cite{binsbergen_thesis}.

The \textbf{FUN-GLL} variant of the GLL algorithm computes the minimal amount of grammar information necessary for generalized parsing~\cite{binsbergen2020}. 
The algorithm can be used by parser combinators because grammar information is computed on an as-needed basis during parsing, rather than in a separate phase.
This way, reuse through abstraction with separate compilation is realized in FUN-GLL, and in our Happy back-end based on FUN-GLL.

\textbf{Parser generators}, viewed as implementing DSLs, typically provide a \emph{fixed} number of language constructs, often corresponding to some variation of Extended Backus-Naur Form (EBNF).
The `parameterized nonterminals\footnote{Referred to as `parameterized productions' in the user manual of Happy.}' of the Happy parser generator make it possible for users to define their own operators over grammar symbols, akin to macro-grammars~\cite{fischer68,thiemann2004} and the parameterized nonterminals of the Menhir~\cite{menhir} and \precc{}~\cite{precc} parser generators.
However, parameterized nonterminals do not reach their full potential in the existing (LALR and GLR) back-ends of Happy as it is not possible to reuse user-defined operators across grammars.
Moreover, the back-ends rely on an algorithm that effectively performs macro-expansion on all parameterized nonterminals.
This algorithm may fail to terminate or may result in exponentially large parsers generated in an exponential amount of time.
The GLL back-end for Happy presented in this paper overcomes these problems by generating reusable, higher-order functions, akin to FUN-GLL parser combinators, for the nonterminals of a grammar.

The contributions of this paper are as follows. This paper:
\begin{itemize}
\item Presents a strategy for generating modular, reusable and complete top-down parsers from syntax descriptions with parameterized nonterminals
\item Adds a back-end to Happy that realizes the full potential of Happy's parameterized nonterminals, making Happy one of the few parser generators to support `reuse through abstraction' of which it is perhaps the first to generate complete parsers that find all derivations of an input string
\item The GLL back-end is to our knowledge the first implementation of GLL in a parser generator for Haskell and the first back-end for Happy with all the benefits of recursive descent parsing
\item The GLL back-end is compared to the existing GLR and LALR back-ends in a number of experiments, demonstrating the characteristics of the new back-end
\end{itemize}
%

Section~\ref{sec_motivation} motivates the GLL back-end by explaining the advantages of recursive descent parsing and parameterized nonterminals.
Section~\ref{sec_fungll} explains how the GLL back-end translates parameterized nonterminals to reusable GLL parsers that compute all interpretations of an input string.
Section~\ref{sec_implementation} discusses practical aspects of the implementation of the GLL back-end, including some specific aspects of the Happy grammar formalism such as monadic semantic actions.
Section~\ref{sec_evaluation} demonstrates certain advantages of the alternative treatment of parameterized nonterminals and compares the running times of the different Happy back-ends in an empirical evaluation. 
Sections~\ref{sec_related_work} and~\ref{sec_conclusion} discuss related work and conclude.

%% file: example_fungll.tex
\begin{table*}[t]
\begin{center}
\newcounter{descr}
\setcounter{descr}{0}
\begin{tabular}{|l|l|l|l|l|l|l|}
\hline
\# & \textbf{processed descriptor} & \textbf{action} & $\mathit{uset}$ \textbf{ext.} & $\mathit{bsrs}$ \textbf{ext.} & $\mathit{grel}$ \textbf{extension} & $\mathit{prel}$ \textbf{ext.}\\
\hline
\descrc &$\descr{E}{}{EEE}{0}{0}$ & \textbf{descend} & \sout{1,2,3} & & $\langle\langle E,0\rangle,\langle\gslot{E}{E}{EE},0\rangle\rangle$&\\
\descrc & $\descr{E}{}{\texttt{a}}{0}{0}$ &\textbf{continue}    & 4 & 5 & &\\
\descrc & $\descr{E}{}{}{0}{0}$ & \textbf{ascend}  & 5 & 1,2 & & $\langle\langle E,0\rangle,0\rangle$\\
\descrc & $\descr{E}{\texttt{a}}{}{0}{1}$ &\textbf{ascend}  & 6 & 6 & & $\langle\langle E,0\rangle,1\rangle$\\
\descrc &$\descr{E}{E}{EE}{0}{0}$ & \textbf{continue}    & 7,8 & 3,7 &$\langle\langle E,0\rangle,\langle\gslot{E}{EE}{E},0\rangle\rangle$ & \\
\descrc &$\descr{E}{E}{EE}{0}{1}$ & \textbf{descend} & 9,10,11 & & $\langle\langle E,1\rangle,\langle \gslot{E}{EE}{E},0\rangle\rangle$ &\\
\descrc &$\descr{E}{EE}{E}{0}{0}$ & \textbf{continue} & 12,13& 4,9 & $\langle\langle E,0\rangle,\langle\gslot{E}{EEE}{},0\rangle\rangle$ &\\
\descrc & $\descr{E}{EE}{E}{0}{1}$ & \textbf{descend} & \sout{9,10,11} & & $\langle\langle E,1\rangle,\langle \gslot{E}{EEE}{},0\rangle\rangle$ &\\
\descrc & $\descr{E}{}{EEE}{1}{1}$ & \textbf{descend} & \sout{9,10,11} & &$\langle\langle E,1\rangle,\langle \gslot{E}{E}{EE},1\rangle\rangle$& \\
\descrc & $\descr{E}{}{\texttt{a}}{1}{1}$ & \sout{\textbf{continue}} & & & & \\
\descrc & $\descr{E}{}{}{1}{1}$ & \textbf{ascend} & \sout{8,13},14 & 8,10,11,12 & & $\langle\langle E,1\rangle,1\rangle$ \\
\descrc & $\descr{E}{EEE}{}{0}{0}$ &\textbf{ascend}& \sout{5,7,12} & \sout{2,3,4} & & \sout{$\langle\langle E,0\rangle,0\rangle$} \\
\descrc & $\descr{E}{EEE}{}{0}{1}$ &\textbf{ascend}& \sout{6,8,13} & \sout{6,7,9} & & \sout{$\langle\langle E,0\rangle,1\rangle$} \\
\descrc & $\descr{E}{E}{EE}{1}{1}$ &\textbf{continue}& 15 & 13 & $\langle\langle E,1\rangle,\langle \gslot{E}{EE}{E}, 1\rangle\rangle$ &\\ 
\descrc & $\descr{E}{EE}{E}{1}{1}$ &\textbf{continue}& 16 & 14 & $\langle\langle E,1\rangle,\langle \gslot{E}{EEE}{}, 1\rangle\rangle$ &\\ 
\descrc & $\descr{E}{EEE}{}{1}{1}$ &\textbf{ascend}& \sout{8,13,14,15,16} & \sout{8,10,12,13,14} & & \sout{$\langle\langle E,1\rangle,1\rangle$}\\ 
\hline
\end{tabular}%
\end{center}%
\caption{Example execution of \fungll{} with $E ::= EEE \;| \;a \;| \;\epsilon$ and string \texttt{"a"}. The BSR elements are given in Figure~\ref{fig_eee_bsr}.}%
\label{tab_run_eee}%
\end{table*}%
\begin{figure}[t]
\begin{align*}
\{&\bsr{\mathbf{E}}{}{}{0}{0}{0}\tag{1}
 ,\\&\bsr{\mathbf{E}}{\mathbf{E}}{\mathbf{E}\;\mathbf{E}}{0}{0}{0}\tag{2}
 ,\\&\bsr{\mathbf{E}}{\mathbf{E}\;\mathbf{E}}{\mathbf{E}}{0}{0}{0}\tag{3}
 ,\\&\bsr{\mathbf{E}}{\mathbf{E}\;\mathbf{E}\;\mathbf{E}}{}{0}{0}{0}\tag{4}
 ,\\&\bsr{\mathbf{E}}{\texttt{a}}{}{0}{0}{1}\tag{5}
 ,\\&\bsr{\mathbf{E}}{\mathbf{E}}{\mathbf{E}\;\mathbf{E}}{0}{0}{1}\tag{6}
 ,\\&\bsr{\mathbf{E}}{\mathbf{E}\;\mathbf{E}}{\mathbf{E}}{0}{0}{1}\tag{7}
 ,\\&\bsr{\mathbf{E}}{\mathbf{E}\;\mathbf{E}}{\mathbf{E}}{0}{1}{1}\tag{8}
 ,\\&\bsr{\mathbf{E}}{\mathbf{E}\;\mathbf{E}\;\mathbf{E}}{}{0}{0}{1}\tag{9}
 ,\\&\bsr{\mathbf{E}}{\mathbf{E}\;\mathbf{E}\;\mathbf{E}}{}{0}{1}{1}\tag{10}
 ,\\&\bsr{\mathbf{E}}{}{}{1}{1}{1}\tag{11}
 ,\\&\bsr{\mathbf{E}}{\mathbf{E}}{\mathbf{E}\;\mathbf{E}}{1}{1}{1}\tag{12}
 ,\\&\bsr{\mathbf{E}}{\mathbf{E}\;\mathbf{E}}{\mathbf{E}}{1}{1}{1}\tag{13}
 ,\\&\bsr{\mathbf{E}}{\mathbf{E}\;\mathbf{E}\;\mathbf{E}}{}{1}{1}{1}\tag{14} \}
\end{align*}%
\caption{BSR set for $E ::= EEE \; | \; a \; | \; \epsilon$ and string \texttt{"a"}.}%
\label{fig_eee_bsr}%
\end{figure}%